\documentclass[10pt,journal,compsoc]{IEEEtran}

\ifCLASSOPTIONcompsoc
  \usepackage[nocompress]{cite}
\else
  \usepackage{cite}
\fi

\ifCLASSINFOpdf
\else
\fi

\usepackage{bm}
\usepackage{amsmath}
\usepackage{multirow}
\usepackage{subfigure}
\usepackage{array}
\usepackage{amsthm}
\usepackage{amsfonts}
\usepackage{graphicx}
\usepackage{rotating}
\usepackage{ragged2e}
\usepackage{color}
\usepackage[linesnumbered,ruled]{algorithm2e}
\usepackage{makecell}

\newcommand{\citet}{\cite}
\newcommand{\figref}[1]{Fig. \ref{#1}}
\newcommand{\tabref}[1]{Table \ref{#1}}
\newcommand{\secref}[1]{Section \ref{#1}}
\newcommand{\appendixref}[1]{the Appendix}
\newcommand{\equref}[1]{Equation (\ref{#1})}
\newcommand{\algoref}[1]{Algorithm \ref{#1}}

\begin{document}
%
\title{
Continuous-Time User Preference Modelling \\ for Temporal Sets Prediction
}
%
%
%

\author{Le~Yu,
        Zihang~Liu,
        Leilei~Sun,
        Bowen~Du,
        Chuanren~Liu,
        Weifeng~Lv
\IEEEcompsocitemizethanks{\IEEEcompsocthanksitem L. Yu, Z. Liu, L. Sun, B. Du and W. Lv are with the State Key Laboratory of Software Development Environment, Beihang University, Beijing, 100191, China.\protect\\
E-mail: \{yule,lzhmark,leileisun,dubowen,lwf\}@buaa.edu.cn
\IEEEcompsocthanksitem C. Liu is with the Department of Business Analytics and Statistics, the University of Tennessee, Knoxville, TN 37996, USA.\protect\\
E-mail: cliu89@utk.edu
}


\thanks{(Corresponding author: Leilei Sun.)}}

%
%

\markboth{IEEE TRANSACTIONS ON KNOWLEDGE AND DATA ENGINEERING,~Vol.~XX, No.~X, XX~XXXX}%
{Yu \MakeLowercase{\textit{et al.}}: Continuous-Time User Preference Modelling for Temporal Sets Prediction}
%

\IEEEtitleabstractindextext{%
\begin{abstract}
\justifying
Given a sequence of sets, where each set has a timestamp and contains an arbitrary number of elements, temporal sets prediction aims to predict the elements in the subsequent set.
Previous studies for temporal sets prediction mainly focus on the modelling of elements and implicitly represent each user’s preference based on his/her interacted elements.
However, user preferences are often continuously evolving and the evolutionary trend cannot be fully captured with the indirect learning paradigm of user preferences.
To this end, we propose a continuous-time user preference modelling framework for temporal sets prediction, which explicitly models the evolving preference of each user by maintaining a memory bank to store the states of all the users and elements.
Specifically, we first construct a universal sequence by arranging all the user-set interactions in a non-descending temporal order, and then chronologically learn from each user-set interaction.
For each interaction, we continuously update the memories of the related user and elements based on their currently encoded messages and past memories. 
Moreover, we present a personalized user behavior learning module to discover user-specific characteristics based on each user's historical sequence, which aggregates the previously interacted elements from dual perspectives according to the user and elements.
Finally, we develop a set-batch algorithm to improve the model efficiency, which can create time-consistent batches in advance and achieve 3.5$\times$ and 3.0$\times$ speedups in the training and evaluation process on average. 
Experiments on four real-world datasets demonstrate the superiority of our approach over state-of-the-arts under both transductive and inductive settings. The good interpretability of our method is also shown.
\end{abstract}

}

\maketitle

\IEEEdisplaynontitleabstractindextext

%
\IEEEpeerreviewmaketitle

\section{Introduction}
\label{section-1}
\IEEEPARstart{T}{emporal} sets can be formalized as a sequence of sets, where each set carries a timestamp and includes an arbitrary number of elements \cite{DBLP:conf/kdd/Benson0T18}. In many practical scenarios, the sequential behaviors of users could be treated as temporal sets. For example, the customers' purchasing behaviors of baskets in online shopping \cite{DBLP:conf/www/RendleFS10,DBLP:conf/sigir/YuLWWT16}, patients' medical behaviors of prescriptions in intelligent treatments \cite{DBLP:conf/ijcai/ZhuLLWGLC17,DBLP:conf/kdd/0001YSLQT18}, and tourists' travelling behaviors of point-of-interests in transportation systems \cite{DBLP:conf/aaai/ZhaoZLXLZSZ19,DBLP:conf/ijcai/ZhangSZLLWKK20}. Accurately predicting which elements will appear in the next-period set can help people make better decisions ahead of time \cite{DBLP:conf/kdd/HuH19,DBLP:conf/sigir/SunBDL0L20,DBLP:conf/kdd/YuSDL0L20,DBLP:conf/www/YuWS0L22}.

\begin{figure}[!htbp]
    \centering
    \includegraphics[width=1.0\columnwidth]{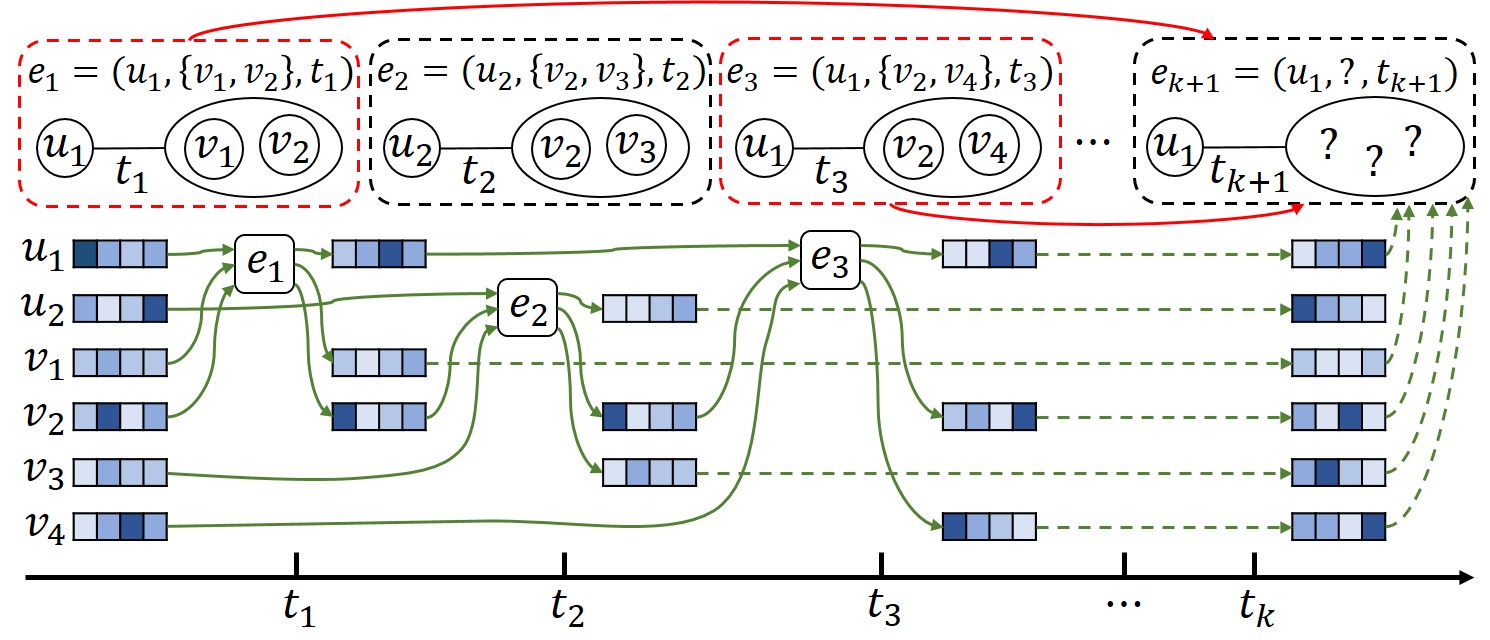}
    \caption{Our task aims to predict the next-period set of user $u_1$ at timestamp $t_{k+1}$. Unlike previous methods that implicitly capture each user's preference based on the interacted elements, our approach maintains a memory bank to \textit{explicitly} store the states of all the users and elements, which could be continuously evolved upon the non-descending sequence of all the user-set interactions (plotted in green). We also learn $u_1$'s personalized behaviors based on his/her historical sequence to reflect the individual characteristics (plotted in red).
    }
    \label{fig:motivation}
\end{figure}

Recently, some researchers have attempted to tackle the problem of temporal sets prediction. One part of the scholars simplified the temporal sets prediction problem by converting it into conventional prediction of temporal events \cite{DBLP:conf/www/LiJTYHC17,DBLP:conf/aaai/ZhaoZLXLZSZ19} through a set embedding process \cite{DBLP:conf/sigir/YuLWWT16,DBLP:conf/kdd/HuH19,DBLP:journals/kbs/ShiYSWLZ21}. Specifically, \citet{DBLP:conf/sigir/YuLWWT16,DBLP:conf/kdd/HuH19} and \citet{DBLP:journals/kbs/ShiYSWLZ21} first derived the representations of sets via pooling operations or matrix factorization, and then learned the temporal dependencies in user behaviors by Recurrent Neural Networks (RNNs) \cite{DBLP:journals/neco/HochreiterS97,DBLP:conf/emnlp/ChoMGBBSB14} and the attention mechanism \cite{DBLP:journals/corr/BahdanauCB14}. The other part of researchers focused on the learning of elements by designing customized components for temporal sets prediction \cite{DBLP:conf/sigir/SunBDL0L20,DBLP:conf/kdd/YuSDL0L20,DBLP:conf/www/YuWS0L22}. In particular, \citet{DBLP:conf/sigir/SunBDL0L20} presented a dual sequential network to learn the multi-level (i.e., set-level and element-level) representation of each user's sequence based on the Transformer architecture \cite{DBLP:conf/nips/VaswaniSPUJGKP17}. \citet{DBLP:conf/kdd/YuSDL0L20} first performed graph convolutions on the co-occurrence graphs of elements in the set level to discover element relationships, and then learned the temporal dependencies and shared patterns of elements by the attention mechanism and the gating mechanism. \citet{DBLP:conf/www/YuWS0L22} first connected the sequences of different users by constructing a temporal graph and then captured high-order user-element interactions via the element-guided message aggregation mechanism, which is aware of the temporal information. 

Although previous methods are insightful for temporal sets prediction, they mainly focus on the modelling of elements and implicitly capture each user's preference according to the interacted elements. However, user preferences are often continuously evolving in real-world scenarios \cite{DBLP:conf/wsdm/ChenXZT0QZ18,DBLP:conf/sigir/HuangZDWC18,DBLP:conf/kdd/KumarZL19,DBLP:conf/sigir/RenQF0ZBZXYZG19}, and thus the evolutionary trend cannot be fully captured by the implicit learning paradigm of user preferences. This motivates us to design a new framework to \textit{explicitly} model the evolving user preferences, see the bottom of \figref{fig:motivation}. Moreover, when making the prediction for each user, it is also essential to discover the personalized behaviors based on the historical sequence for better reflecting the user-specific characteristics, see the top of \figref{fig:motivation}.

In this paper, we propose a \underline{\textbf{C}}ontinuous-\underline{\textbf{T}}ime user preference modelling framework for \underline{\textbf{T}}emporal \underline{\textbf{S}}ets \underline{\textbf{P}}rediction, namely CTTSP. Our main idea is to explicitly capture the evolving user preferences in a continuous manner by storing the states of all the users and elements via a memory bank. Given the sequences of all the users, we first sort all the user-set interactions in a non-descending temporal order to construct a universal sequence, and then chronologically learn from each user-set interaction.
In particular, for each user-set interaction, our approach first learns the relationships of the related user and elements via message encoders, and then updates their memories based on the encoded messages and previous memories via memory updaters.
To exploit each user's unique characteristics, we also devise a personalized user behavior learning component, which adaptively aggregates the historical sequence with the guidance of the user and elements from dual perspectives.
Finally, a set-batch algorithm is developed to make our method more efficient, which can create time-consistent batches in advance and achieve the average of 3.5$\times$ and 3.0$\times$ speedups in training and evaluation, respectively.
Experiments on four benchmarks not only show that our approach could significantly outperform existing methods under both transductive and inductive settings, but also reveal the good interpretability of our method.
Our key contributions include:
\begin{itemize}
    \item
    We present a \textit{continuous-time user preference modelling framework to explicitly capture the evolving user preferences} by maintaining a memory bank to store the states of all the users and elements. When a user-set interaction is observed, we continuously update the related user's and elements' memories using the encoded messages and past memories. 
    
    \item
    We learn the \textit{personalized behaviors of each user from the historical sequence} by adaptively aggregating the interacted elements based on the user and elements.
    
    \item
    \textit{A set-batch algorithm is developed to improve the model efficiency}, which can create time-consistent batches and accelerate the model speed in the training and evaluation process.
\end{itemize}

The key difference between previous \textit{implicit} methods and our \textit{explicit} approach is whether the model can \textit{memorize users’ historical states and track the evolving user preferences across time}. Previous methods do not design customized modules to maintain the states of users and they implicitly represent user preferences based on the interacted elements. For each user-set interaction, they need to recompute the user representation from scratch since they lack the ability in memorizing users’ historical states. Instead, our approach uses a memory bank to explicitly store the states of users and elements, which can be updated to be the latest based on the currently encoded messages and the memorized state of the user. The explicit learning paradigm helps our approach capture the continuously evolving user preferences, which cannot be achieved by previous methods. 

We would like to point out that there is another understanding that a \textit{continuous-time} model should be able to derive user preferences at any given time, which is the objective of methods like Neural Ordinary/Partial Differential Equations \cite{DBLP:conf/nips/ChenRBD18,DBLP:conf/iclr/LiKALBSA21}. Although insightful, these methods tend to be computationally expensive, especially when using black-box differential equation solvers. Their performance is also sensitive to the choice of numerical solvers. Furthermore, our approach learns from the universal sequence constructed by all the user-set interactions, which differs from the input data format of the above methods. Hence, we conclude that it is challenging to design such a continuous-time method for temporal sets prediction, which could be left as an interesting problem for future research.

\section{Related work}
\label{section-2}

\subsection{Temporal Sets Prediction}
Over the past decades, time series \cite{brockwell2016introduction,DBLP:conf/aaai/ZhouZPZLXZ21} and temporal events \cite{DBLP:conf/www/LiJTYHC17,DBLP:conf/aaai/ZhaoZLXLZSZ19} have been widely studied in the temporal data mining community. However, as a more complicated type of temporal data, temporal sets have not been fully investigated yet.
Formally, temporal sets could be formalized as a sequence of sets, where each set contains an arbitrary number of elements and is associated with a timestamp \cite{DBLP:conf/kdd/Benson0T18}. 
In real-world scenarios, temporal sets are ubiquitous and accurately predicting of temporal sets can help people make better decisions in advance \cite{DBLP:conf/kdd/HuH19,DBLP:conf/sigir/SunBDL0L20,DBLP:conf/kdd/YuSDL0L20,DBLP:conf/www/YuWS0L22}.

Indeed, the prediction of temporal sets is much more challenging than the predictive modelling of time series and temporal events. The existing methods for temporal sets prediction could be divided into the following two categories. 
The first category of methods adopted a set embedding phase to denote each set as a vectorized representation, and then fed the sequence of set representations into sequence models to learn the temporal dependencies in user behaviors. For instance, \citet{DBLP:conf/sigir/YuLWWT16} adopted pooling operations to get the fixed-length representation of each basket and leveraged RNNs to learn the dynamic patterns in each customer's purchasing behaviors. \citet{DBLP:conf/kdd/HuH19} and \citet{DBLP:journals/kbs/ShiYSWLZ21} first obtained each set's representation by the average pooling or matrix factorization, and then employed RNNs with the attention mechanism to capture the temporal dependencies. The repeated elements in each user's sequence are also considered. 
Methods belonging to the second category designed specialized modules to investigate the elements in sets for further improvements. \citet{DBLP:conf/sigir/SunBDL0L20} proposed a dual sequential network with the co-transformer architecture to compute both set-level and element-level representations of each user's sequence. \citet{DBLP:conf/kdd/YuSDL0L20} first learned the relationships of elements within each set based on the constructed set-level co-occurrence graphs, and then extended the attention mechanism and the gating mechanism to mine the temporal dependencies and shared patterns among elements. \citet{DBLP:conf/www/YuWS0L22} pointed out the necessity of leveraging the collaborative signals for temporal sets prediction and learned element-specific representations for each user on the constructed temporal graph with the usage of temporal information.

Existing methods for temporal sets prediction implicitly capture the user preference according to the interacted elements and thus fail to capture the continuous evolutionary trend of user preference. To cope with this issue, we present a new framework that explicitly models the evolving dynamics in user preference in a continuous-time manner.

\subsection{User Modelling}
Based on the massive collected data of users, a great number of efforts have been made on user modelling in recent years. Some methods utilized users' sequences to exploit their dynamics and formalized the task as a sequence learning problem. For example, \citet{DBLP:conf/www/RendleFS10} integrated the Markov chain and matrix factorization to learn the evolving preferences of users. \citet{DBLP:journals/corr/HidasiKBT15} proposed GRU4Rec based on RNNs for session-based recommendation. \citet{DBLP:conf/wsdm/ChenXZT0QZ18} designed the memory-augmented neural network to capture user sequential behaviors for sequential recommendation. \citet{DBLP:conf/sigir/HuangZDWC18} incorporated the knowledge base information into RNN-based memory networks to enhance the model interpretability. \citet{DBLP:conf/sigir/RenQF0ZBZXYZG19} proposed a lifelong sequential modelling framework to learn the multi-scale evolving patterns in user behaviors via a hierarchical and periodical updating mechanism. There are also some attempts for user modelling in a non-chronological manner. For click-through rate prediction, \citet{DBLP:conf/kdd/ZhouZSFZMYJLG18} computed user representations that are specific to the ads by a local activation unit. \citet{DBLP:conf/kdd/YingHCEHL18} and \citet{DBLP:conf/sigir/Wang0WFC19} extended the Graph Neural Networks (GNNs) \cite{DBLP:conf/iclr/KipfW17,DBLP:conf/nips/HamiltonYL17,DBLP:conf/iclr/VelickovicCCRLB18} to recommender systems by learning from the user-item interaction graph. They performed graph convolutions to generate embeddings of both users and items, which utilize the graph structure and node feature information simultaneously. \citet{DBLP:conf/sigir/0001DWLZ020} removed the feature transformation and nonlinear activation in graph convolutions, and linearly propagated information on the user-item interaction graph. 

In this paper, we design a continuous-time user preference modelling framework for temporal sets prediction.

\subsection{Continuous-Time Dynamic Graph Learning}
A continuous-time dynamic graph is defined as a chronological sequence of events, where each event can correspond to the modifications of an object or the interactions of multiple objects \cite{DBLP:journals/jmlr/KazemiGJKSFP20}. Representation learning on continuous-time dynamic graphs aims to provide meaningful representations for nodes in the graph with the consideration of graph dynamics. 
\citet{DBLP:conf/cikm/ChangLW0FS020} first constructed temporal dependency interaction graphs which are induced from the sequence of interactions, and then captured both global and local graph information with a dynamic message passing neural network. \citet{DBLP:conf/iclr/XuRKKA20} first considered the temporal dependencies via a time encoding function and then aggregated the neighbor information using the temporal graph attention layer to compute time-aware node representations. \citet{DBLP:journals/corr/abs-2006-10637} designed a general framework for continuous-time dynamic graphs by learning evolving node representations across time. To learn from temporal user-item interaction graphs, \citet{DBLP:conf/kdd/KumarZL19} employed two RNNs to update the embedding of a user and an item and introduced a projection operator to learn the embedding trajectories of users and items. \citet{DBLP:conf/cikm/FanLZX0Y21} defined a continuous-time bipartite graph of users and items, and designed a temporal collaborative transformer to consider the temporal dynamics inside sequential patterns and mine the underlying collaborative signals. 
\citet{yu2023towards} introduced a unified library for facilitating the development of the continuous-time dynamic graph learning field.

However, the above methods are not specifically designed for predicting temporal sets. They fail to capture the relationships between elements within the same set as well as the correlations between the user and the interacted set.
In this paper, we propose a customized continuous-time learning framework for temporal sets prediction.

\section{Problem Formalization}
\label{section-3}


Let $\mathbb{U}=\{u_1,\cdots,u_m\}$ and $\mathbb{V}=\{v_1,\cdots,v_n\}$ denote the collections of $m$ users and $n$ elements. We use $\mathcal{S}^t=\{e_1,e_2,\cdots,e_K\}$ to represent the observed user-set interactions in chronological order, where $e_k=(u^e_k,\mathbb{S}_k,t_k)$ is an interaction event, $u^e_k\in \mathbb{U}$, $\mathbb{S}_k \subset \mathbb{V}$, and $0 < t_1 \leq \cdots \leq t_K \leq t$. We use $T_i$ to represent the timestamp of user $u_i$'s last interaction set.
The task of \textbf{temporal sets prediction} aims to predict elements to appear in the subsequent interaction set for each user $u_i \in \mathbb{U}$ at the next-period timestamp $T_i^\prime$. Our learning paradigm for user $u_i$ could be represented by
$$\hat{\mathbb{S}}_i^{T_i^\prime} = f\left(\mathcal{S}^{T_i}\right).$$
It is worth noticing that our solution integrates the historical behaviors of all the users until timestamp $T_i$ when making predictions for user $u_i$, who can benefit from exploiting the collaborative signals. 

In this paper, compared with the existing methods that implicitly represent each user's preference based on his/her interacted elements, we aim to design a new framework to explicitly capture the evolving user preference by learning the continuous-time representations of users and elements.


\section{Methodology}
\label{section-4}
This section shows the framework of our CTTSP and presents each component one by one.
\begin{figure*}[!htbp]
    \centering
    \includegraphics[width = 2.00\columnwidth]{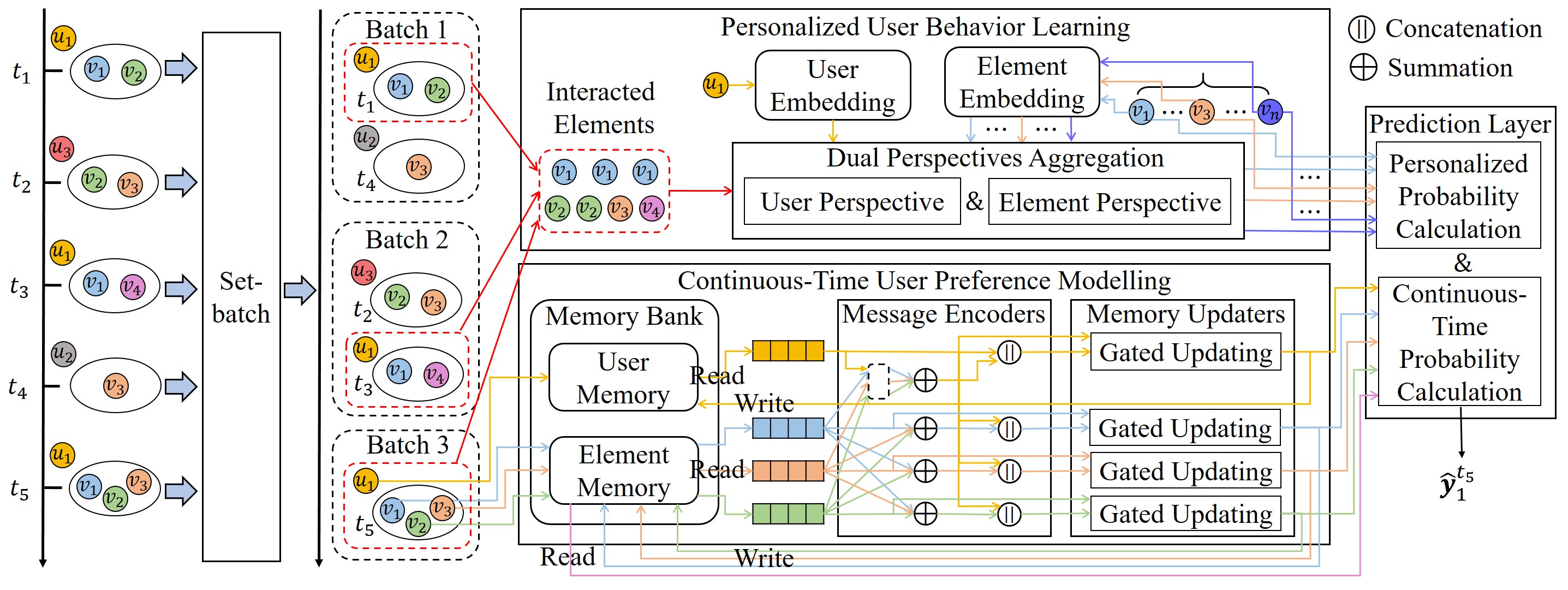} 
    \caption{Framework of the proposed model.}
    \label{fig:framework}
\end{figure*}
As shown in \figref{fig:framework}, our approach predicts the next-period set for each user in a continuous manner. To explicitly capture the time-evolving user preferences, we maintain a memory bank for storing the states of all the users and elements. We first construct a universal sequence by arranging all the user-set interactions in a non-descending temporal order, and then learn from each user-set interaction chronologically. For each user-set interaction, we learn the relationships between the user and elements by message encoders, and then update their memories to be the latest by memory updaters using the encoded messages and past memories. We also devise a personalized user behavior learning module to exploit each user's individual characteristics from dual perspectives, which adaptively aggregates elements in the historical sequence with the guidance of the user and elements. A prediction layer is designed to compute the appearing probabilities of elements in the next-period set by integrating the continuous-time and personalized information. Moreover, we present a set-batch algorithm to improve the model efficiency by creating time-consistent batches in advance with the constraint of temporal orders.

\subsection{Continuous-Time User Preference Modelling}
We first construct a universal sequence by sorting all the user-set interactions in a non-descending order based on the timestamp of each set (as shown on the left of \figref{fig:framework}), and then learn from user-set interactions chronologically for encoding the collaborative signals among different users. To be specific, we maintain a memory bank to explicitly store the states of all the users and elements for tracking the evolving preferences. We initialize the memories of all the users and elements as zero vectors, and update the memories of the related user and elements involved in each user-set interaction according to the following message encoders and memory updaters.

\subsubsection{Message Encoders}
In a user-set interaction, the involved user and elements are often associated with each other, including the mutual correlations between the user and elements, and the co-occurring relations of elements. Therefore, we design two customized message encoders to learn the above user-element and element-element relationships. 


\textbf{User Message Encoder.}
Mathematically, We use $\bm{z}_{i}^{u,t}, \bm{z}_{j}^{v,t} \in \mathbb{R}^{d}$ to denote the memories of user $u_i$ and element $v_j$ at timestamp $t$, where $d$ is the hidden dimension. When a user-set interaction $e_k=(u^e_k,\mathbb{S}_k,t_k)$ is observed, the correlation weight $\alpha_{i,j}^{k}$ of the user $u_i=u^e_k$\footnote{For better readability, we assume that $u^e_k$ corresponds to the $i$-th user $u_i \in \mathbb{U}$.} and elements $v_j \in \mathbb{S}_k$ are calculated by the following attention mechanism
\begin{equation}
    \label{equ:user_message_encoder_att}
    \alpha_{i,j}^{k} = \mathop{softmax}\limits_{\forall v_j \in \mathbb{S}_k}\left({\left(\bm{W}_Q^u \bm{z}_i^{u,t_k^-(u_i)}\right)^\top \left(\bm{W}_K^u \bm{z}_j^{v,t_k^-(v_j)}\right)}/{\sqrt{d}}\right),
\end{equation}
where $\bm{W}_Q^u, \bm{W}_K^u \in \mathbb{R}^{d \times d}$ are the query and key vector of users. $t_k^-(u_i)$ and $t_k^-(v_j)$ are the last time that user $u_i$ interacted with a set and element $v_j$ was involved in a set right before timestamp $t_k$, respectively. 
Then, we aggregate the elements in $\mathbb{S}_k$ for user $u_i$ using the learned weights by
\begin{equation}
    \label{equ:user_message_encoder_aggregation}
    \bm{c}_i^{u,t_k} = \sum_{v_j \in \mathbb{S}_k} \alpha_{i,j}^{k} \cdot \bm{W}_V^u \bm{z}_j^{v,t_k^-(v_j)},
\end{equation}
where $\bm{W}_V^u \in \mathbb{R}^{d \times d}$ is the value vector of users. $\bm{c}_i^{u,t_k}$ adaptively selects the information of all the elements in $\mathbb{S}_k$, and thus captures the user-element relationships. 
Finally, the message of user $u_i$ at timestamp $t_k$ is obtained by concatenating $\bm{c}_i^{u,t_k}$ and $\bm{z}_i^{u,t_k^-(u_i)}$, that is,
\begin{equation}
    \label{equ:user_message_encoder_concat}
    \bm{m}_i^{u,t_k} = \bm{c}_i^{u,t_k} \| \bm{z}_i^{u,t_k^-(u_i)}.
\end{equation}

\textbf{Element Message Encoder.}
The calculation process of the element message encoder is analogous to that of the user message encoder, and the main difference is that we need to additionally design trainable parameters specific to elements, including $\bm{W}_Q^v, \bm{W}_K^v, \bm{W}_V^v \in \mathbb{R}^{d \times d}$. Therefore, we do not elaborate on them in detail. Based on the element message encoder, we could obtain $\bm{m}_j^{v,t_k}$, which is the message of element $v_j \in \mathbb{S}_k$ at timestamp $t_k$.

\subsubsection{Memory Updaters}
Our approach maintains a memory bank to store the memories of both users and elements, where each memory aims to represent the historical states of the corresponding user or element in a compact format. Note that the memories are \textit{not trainable}, but they can be modified according to the trainable parameters in the following memory updaters. Each memory will be continuously updated when a new user-set interaction contains the user or the element.

\textbf{User Memory Updater.}
For the user-set interaction $e_k$, we keep the memory of user $u_i$ to be the latest by the gated updating mechanism with the consideration of both encoded message and previous memory as follows,
\begin{equation}
    \label{equ:user_memory_updater}
    \resizebox{1.0\hsize}{!}{
    $\bm{z}_i^{u,t_k} = tanh\left(\bm{g}_{i,M}^{u,t_k} \odot \bm{W}_M^u \bm{m}_i^{u,t_k} + (\bm{1} - \bm{g}_{i,M}^{u,t_k}) \odot \bm{W}_Z^u \bm{z}_i^{u,t_k^-(u_i)}\right)$},
\end{equation}
where $\bm{W}_M^u \in \mathbb{R}^{d \times 2d}$ and $\bm{W}_Z^u \in \mathbb{R}^{d \times d}$ are the transformation matrices to align the dimensions of each user's message and memory. $\odot$ denotes the Hadamard product. $\bm{g}_{i,M}^{u,t_k} \in \mathbb{R}^d$ controls the importance of the user message $\bm{m}_i^{u,t_k}$, which is computed by
\begin{equation}
    \label{equ:user_memory_updater_gated}
    \bm{g}_{i,M}^{u,t_k} = \frac{\exp{\left(\bm{M}^u \bm{W}_M^u \bm{m}_i^{u,t_k}\right)}}{\exp{\left(\bm{M}^u \bm{W}_M^u \bm{m}_i^{u,t_k}\right)} + \exp{\left(\bm{Z}^u \bm{W}_Z^u \bm{z}_i^{u,t_k^-(u_i)}\right)}},
\end{equation}
where $\bm{M}^u, \bm{Z}^u \in \mathbb{R}^{d \times d}$ are user-specific trainable parameters to calculate the importance of each dimension of the message and memory. 

\textbf{Element Memory Updater.}
The computations of the element memory updater are similar to those of the user memory updater. One difference is that we need to design element-specific trainable parameters, including $\bm{W}_M^v \in \mathbb{R}^{d \times 2d}, \bm{W}_Z^v, \bm{M}^v, \bm{Z}^v \in \mathbb{R}^{d \times d}$. Hence, we do not elaborate on them due to space limitations. Based on the element memory updater, we could obtain the memory of element $v_j \in \mathbb{S}_k$ at timestamp $t_k$, which is denoted by $\bm{z}_j^{v,t_k}$.

Note that our method learns from the universal sequence which is composed of all the user-set interactions, and the memories of users and elements are continuously updated for each user-set interaction. This provides our approach with the ability in \textit{exploiting the underlying collaborative signals across different users}. Take the first two user-set interactions $e_1=(u_1,\{v_1,v_2\},t_1)$ and $e_2=(u_3,\{v_2,v_3\},t_2)$ in \figref{fig:framework} as an example, after dealing with $e_1$, element $v_2$'s memory will contain the information of element $v_1$ and user $u_1$. Therefore, when learning on $e_2$, user $u_3$ is able to additionally access the cross-user information of user $u_1$ and element $v_1$ via the memory of element $v_2$, and thus the latent collaborative signals could be learned.

\subsection{Personalized User Behavior Learning}
Apart from learning from the universal sequence, we also explore the personalized behaviors of each user via the previously interacted elements from his/her own historical sequence. In particular, for user $u_i$ that involved in the interaction $e_k$, we use $\mathcal{V}_i^{t_k}=\left\{{v_j | \ v_j \in \mathbb{S}_{k^\prime} \wedge u^e_{k^\prime}=u_i \wedge 0 < t_{k^\prime} \leq t_k}\right\}$ to denote the list of elements that $u_i$ have interacted with up to timestamp $t_k$. It is worth noticing that $\mathcal{V}_i^{t_k}$ could contain the same element multiple times, and elements that appear with higher times indicate that they are more preferred by user $u_i$. 
Then, we adaptively aggregate the elements in $\mathcal{V}_i^{t_k}$ from dual perspectives, i.e., the user perspective and the element perspective.

\subsubsection{User Perspective}\label{sec:user_pespective_sec}
Formally, we use $\bm{e}^u, \bm{e}_j^v \in \mathbb{R}^d$ to denote the static embeddings of user $u_i$ and element $v_j$\footnote{Note that $\bm{e}^u$ is shared across users while $\bm{e}_j^v$ is element-specific. 
This can reduce the model complexity and provide our model with the inductive ability (validated in the experiments).}, which are randomly initialized from a normal distribution with mean 0 and variance 1 and can be optimized in the model training process. Then, we use $\bm{e}^u$ as a query vector to compute the correlation of each element $v_j \in \mathcal{V}_i^{t_k}$ by
\begin{equation}
    \label{equ:user_perspective_att}
    \beta_{i,j}^{k} = \frac{\exp{\left(\sigma\left(s\left(\bm{e}^u,\bm{e}_j^v\right)\right)\right)}}{\sum_{v_{j^\prime} \in \mathcal{V}_i^{t_k}} \exp{\left(\sigma\left(s\left(\bm{e}^u,\bm{e}_{j^\prime}^v\right)\right)\right)}},
\end{equation}
where $s(\bm{x},\bm{y})=\bm{x}^\top \bm{y}$ denotes the dot product for calculating the similarity of $\bm{x}$ and $\bm{y}$. $\sigma(\cdot)$ is the $LeakyReLU$ activation function. $\beta_{i,j}^{k}$ represents the importance of element $v_j$ to user $u_i$. Through the above aggregation, elements that are more similar to user $u_i$ will be assigned with higher weights and contribute more in the aggregation process.

\subsubsection{Element Perspective}
We also use each element $v_{j^\prime} \in \mathbb{V}$ as a query vector to calculate the correlation of every element $v_j \in \mathcal{V}_i^{t_k}$ via
\begin{equation}
    \label{equ:element_perspective_att}
    \gamma_{j^\prime,j}^{k} = \frac{\exp{\left(\sigma\left(s\left(\bm{e}_{j^\prime}^v,\bm{e}_j^v\right)\right)\right)}}{\sum_{v_{j^{\prime \prime}} \in \mathcal{V}_i^{t_k}} \exp{\left(\sigma\left(s\left(\bm{e}_{j^\prime}^v,\bm{e}_{j^{\prime \prime}}^v\right)\right)\right)}}.
\end{equation}
$\gamma_{j^\prime,j}^{k}$ stands for the relevance of element $v_{j^\prime}$ to element $v_j$, and elements that are more similar to element $v_{j^\prime}$ will become more important.

\subsubsection{Dual Perspectives Aggregation} 
After obtaining the relationships of each element $v_j \in \mathcal{V}_i^{t_k}$ to user $u_i$ and element $v_{j^\prime} \in \mathbb{V}$, we then make aggregation from the dual perspectives via
\begin{equation}
    \label{equ:dual_perspective_aggregation}
    \bm{h}_{i,j^\prime}^{t_k} = \sum_{v_j \in \mathcal{V}_i^{t_k}} \left(\lambda_{up} \beta_{i,j}^{k} + \left( 1 - \lambda_{up} \right) \gamma_{j^\prime,j}^{k} \right) \cdot \bm{e}_j^v,
\end{equation}
where $0 \leq \lambda_{up} \leq 1$ is a hyperparameter to control the importance of user perspective. $\bm{h}_{i,j^\prime}^{t_k}$ is the aggregation of user $u_i$'s interacted elements until timestamp $t_k$, which is guided by the embeddings of user $u_i$ and element $v_{j^\prime}$.

\subsection{The Prediction Layer}
To provide the appearing probabilities of all the elements in the subsequent set of each user, we design a prediction layer by integrating continuous-time and personalized information. Specifically, the prediction layer first computes the continuous-time and personalized probabilities and then fuses them to make the final predictions.

\subsubsection{Continuous-Time Probability Calculation}
Through the modelling of continuous-time user preferences, for user-set interaction $e_k$, we can obtain the updated memories of the related user $u_i$ and element $v_j \in \mathbb{S}_k$, i.e., $\bm{z}_i^{u,t_k}$ and $\bm{z}_j^{v,t_k}$. We can also get the elements that user $u_i$ have previously interacted with but do not appear in $\mathbb{S}_k$, which belong to $unique(\mathcal{V}_i^{t_k}) \setminus \mathbb{S}_k$. $unique(\cdot)$ is a function to derive the unique items in the input. We denote the memories of element $v_j \in unique(\mathcal{V}_i^{t_k}) \setminus \mathbb{S}_k$ before timestamp $t_k$ as $\bm{z}_j^{v,t_k^-(v_j)}$.
Then, the continuous-time probabilities of elements in the above two parts are calculated as follows.

For element $v_j \in \mathbb{S}_k$, we compute the continuous-time probability that it will appear in the next-period set by the dot product of $\bm{z}_i^{u,t_k}$ and $\bm{z}_j^{v,t_k}$ as follows, 
\begin{equation}
    \label{equ:evolutionary_prob_current}
    \hat{p}_{i,j}^{c,t_k} = s\left(\bm{z}_i^{u,t_k}, \bm{z}_j^{v,t_k}\right).
\end{equation}
For element $v_j \in unique(\mathcal{V}_i^{t_k}) \setminus \mathbb{S}_k$, we should notice that its memory $\bm{z}_j^{v,t_k^-(v_j)}$ is not trainable, and directly using it to calculate the dot product with $\bm{z}_i^{u,t_k}$ will make it fail to access the gradient. To tackle this issue, we first use a single-layer fully connected network (denoted as $FCN(\cdot)$) to project $\bm{z}_j^{v,t_k^-(v_j)}$, and then calculate the dot product of $\bm{z}_i^{u,t_k}$ and the projected memory as follows,
\begin{equation}
    \label{equ:evolutionary_prob_historical}
    \hat{p}_{i,j}^{c,t_k} = s\left(\bm{z}_i^{u,t_k}, FCN\left(\bm{z}_j^{v,t_k^-(v_j)}\right)\right).
\end{equation}

\subsubsection{Personalized Probability Calculation}
According to the personalized user behavior learning component, up to the timestamp $t_k$, we can get $\bm{h}_{i,j^\prime}^{t_k}$ for each $v_{j^\prime} \in \mathbb{V}$ via \equref{equ:dual_perspective_aggregation}, which stands for the aggregation of user $u_i$'s historically interacted elements until timestamp $t_k$ with the guidance of user $u_i$ and element $v_{j^\prime}$. The personalized probability that element $v_{j^\prime}$ will appear in the next-period set is calculated via
\begin{equation}
    \label{equ:personalized_prob}
    \hat{p}_{i,j^\prime}^{s,t_k} = s\left(\bm{W}_S\bm{h}_{i,j^\prime}^{t_k}, \bm{e}_{j^\prime}^{v}\right),
\end{equation}
where $\bm{W}_S \in \mathbb{R}^{d \times d}$ stands for a transformation matrix.

\subsubsection{Probabilities Fusion}
We predict the next-period set by fusing both the continuous-time and personalized probabilities. Specifically, we utilize a hyperparameter $0 \leq \lambda_{cp} \leq 1$ to control the importance of the continuous-time probability by
\begin{equation}
    \label{equ:predict_prob}
    \hat{y}_{i,j}^{t_k} = sigmoid \left(\tau_{i,j}^{t_k} \cdot \lambda_{cp} \cdot \hat{p}_{i,j}^{c,t_k} + (1 - \tau_{i,j}^{t_k} \cdot \lambda_{cp}) \cdot \hat{p}_{i,j}^{s,t_k}\right),
\end{equation}
where $\hat{y}_{i,j}^{t_k}$ stands for the probability that element $v_j$ will appear in the next-period set of user $u_i$ after timestamp $t_k$. $\bm{\tau}_i^{t_k} \in \mathbb{R}^n$ is an indicator vector that is composed of 0 or 1, where the entry of 1 means the corresponding element has interacted with user $u_i$ up to timestamp $t_k$. Otherwise, the entries are set to 0.
We can observe that for element $v_j$ that has interacted with user $u_i$ up to timestamp $t_k$, the appearing probability comes from both the continuous-time probability $\hat{p}_{i,j}^{c,t_k}$ and personalized probability $\hat{p}_{i,j}^{s,t_k}$. For element $v_j$ that has not interacted with user $u_i$ up to timestamp $t_k$, the appearing probability only depends on the personalized probability $\hat{p}_{i,j}^{s,t_k}$.

\subsection{Set-batch Algorithm}
Our approach constructs a universal sequence in the non-decreasing temporal order and processes each user-set interaction one after another, which makes it time-consuming when handling datasets with long time spans. Classical RNN models assume that different users are independent of each other, so these methods can split the sequences of users into multiple batches and compute on each batch in parallel \cite{DBLP:journals/corr/HidasiKBT15,DBLP:conf/kdd/BaytasXZWJZ17}. However, our approach merges the sequences of users into a universal sequence and thus the users are correlated with each other, making the above technique infeasible. Though \citet{DBLP:conf/kdd/KumarZL19} proposed an algorithm to accelerate the model efficiency on temporal interaction networks, it can only handle a pairwise user-element interaction and thus fails to handle the user-set interactions since each set contains an arbitrary number of elements. 

To this end, we develop a set-batch algorithm to create time-consistent batches in advance and improve our model efficiency. The core of the set-batch algorithm is to satisfy the constraint of temporal orders, such that the memory of any user or any element is updated chronologically. Formally, for $\forall u_i \in \mathbb{U}, v_j \in \mathbb{V}, e_k=(u^e_k,\mathbb{S}_k,t_k), e_{k^\prime}=(u^e_{k^\prime},\mathbb{S}_{k^\prime},t_{k^\prime})$, if $u_i=u^e_k \wedge u_i=u^e_{k^\prime} \wedge t_k < t_{k^\prime}$, then $BatchID(e_k) < BatchID(e_{k^\prime})$. Moreover, if $v_j \in \mathbb{S}_k \wedge v_j \in \mathbb{S}_{k^\prime} \wedge t_k < t_{k^\prime}$, then $BatchID(e_k) < BatchID(e_{k^\prime})$. $BatchID(e_k)$ is a function to return the index of the batch that contains $e_k$. 

We show the set-batch algorithm in \algoref{alg:set_batch}. In particular, the set-batch algorithm takes the universal non-decreasing sequence $\mathcal{S}^t=\{e_1,\cdots,e_K\}$ as the input and provides the batched data $\mathcal{B}=\left\{B_1,\cdots,B_{BatchNum}\right\}\footnote{For better understanding, we prescribe that the indices of a list start with 1 rather than 0.}$ as the output, where $BatchNum$ denotes the number of batches. Set-batch first initializes $\mathcal{B}$, $user\_idx\_dict$ and $element\_idx\_dict$ in line 1 and line 2, where $user\_idx\_dict$ and $element\_idx\_dict$ are designed to record the largest index of the batch that contains users and elements, respectively. Then, set-batch chronologically processes each user-set interaction $e_k$ in line 3. It obtains the index of batch that $e_k$ should be inserted from line 4 to line 18 and appends $e_k$ to the corresponding position from line 19 to line 22. Finally, the indices of the related user $u^e_k$ and elements $v_j \in \mathbb{S}_k$ in $user\_idx\_dict$ and $element\_idx\_dict$ are updated from line 23 to line 26. An example of the set-batch algorithm is shown on the left of \figref{fig:framework}.

\begin{algorithm}
\SetKwInOut{Input}{Input}
\SetKwInOut{Output}{Output}
\SetKwComment{Comment}{/* }{ */}
\caption{Set-batch algorithm}
\label{alg:set_batch}
\Input{A sequence of user-set interactions $\mathcal{S}^t=\{e_1,\cdots,e_K\}$ with $0 < t_1 \leq \cdots \leq t_K \leq t$\;}
\Output{The batched data $\mathcal{B}=\left\{B_1,\cdots,B_{BatchNum}\right\}$\;}
Initialize $\mathcal{B}$ to be an empty list\;
Initialize $user\_idx\_dict$ and $element\_idx\_dict$ to be empty dictionaries\;
\For{user-set interaction $e_k=(u^e_k,\mathbb{S}_k,t_k) \in \mathcal{S}^t$}{
    \eIf{$u^e_k$ not in $user\_idx\_dict$}{
        user\_idx $\gets$ 0\;
    }{
        user\_idx $\gets$ $user\_idx\_dict$[$u^e_k$]\;
    }
    element\_max\_idx $\gets$ -1\;
    \For{element $v_j \in \mathbb{S}_k$}{
        \eIf{$v_j$ not in $element\_idx\_dict$}{
            element\_idx $\gets$ 0\;
        }{
            element\_idx $\gets$ $element\_idx\_dict$[$v_j$]\;
        }
        element\_max\_idx $\gets$ max(element\_max\_idx, element\_idx)\;
    }
    insert\_idx $\gets$ max(user\_idx, element\_max\_idx) + 1\;
    \If{insert\_idx \textgreater \ len($\mathcal{B}$)
    }{
        Append an empty list to $\mathcal{B}$\;
    }
    Append $e_k$ to $\mathcal{B}$[insert\_idx]\;
    $user\_idx\_dict$[$u^e_k$] $\gets$ insert\_idx\;
    \For{element $v_j \in \mathbb{S}_k$}{
        $element\_idx\_dict$[$v_j$] $\gets$ insert\_idx\;
    }
}
\end{algorithm}

The complexity of our set-batch algorithm is $\mathcal{O}(|\mathcal{S}^t|)$, which is linear to the number of user-set interactions in the sequence $\mathcal{S}^t$. We do not need to manually predetermine $BatchNum$ because set-batch can automatically provide $BatchNum$ according to different datasets. The value of $BatchNum$ ranges from 1 to $|\mathcal{S}^t|$. The extreme cases are: 1) all the user-set interactions have unique users and elements, and then $BatchNum$ is equal to 1; 2) all the user-set interactions have the same user or the same element, then $BatchNum$ is set to $|\mathcal{S}^t|$.
It can be verified that the set-batch algorithm satisfies the constraint of temporal orders because it guarantees: 1) each user and each element should appear at most once in each batch; and 2) if batch $B_{b}$ and $B_{b^\prime}$ consist of interactions $e_k$ and $e_{k^\prime}$ with $t_k < t_{k^\prime}$, and $e_k$ and $e_{k^\prime}$ contain the same user or the same element, then $b < b^\prime$.

\subsection{Model Training Process}
To support the calculation on the batched data $\mathcal{B}$, we first perform the padding operation in each batch and then use the masking technique in \citet{DBLP:conf/nips/VaswaniSPUJGKP17} by setting all the padded values to -$\infty$ before the softmax operations. 
We treat the task of temporal sets prediction as a multi-label classification problem, where an element corresponds to a label. The ground truth of user $u_i$'s next-period set after timestamp $t_k$ is denoted by $\bm{y}_{i}^{t_k} \in \left\{0,1\right\}^n$, where the entry of 1 indicates the corresponding element appears in the next-period set of user $u_i$. The multi-label classification problem is converted into multiple binary classification problems, which are optimized by minimizing the cross-entropy loss,
\begin{equation}
    \label{equ:loss_function}
    L=-\sum_{e_k \in \mathcal{S}^t \wedge u_i=u^e_k}\sum_{v_j \in \mathbb{V}} {y_{i,j}^{t_k}\log(\hat{y}_{i,j}^{t_k}) + (1-y_{i,j}^{t_k})\log(1-\hat{y}_{i,j}^{t_k})},
\end{equation}
where $y_{i,j}^{t_k}$ and $\hat{y}_{i,j}^{t_k}$ denote the ground truth and predicted probability of element $v_j$ appearing in the next-period set of user $u_i$ after timestamp $t_k$. 
We have also tried other widely-used loss functions in recommender systems such as Sampled Softmax \cite{DBLP:journals/corr/abs-2201-02327} and Personalized Ranking with Importance Sampling \cite{DBLP:conf/www/Lian0C20}, but they cannot show better performance. 
In \appendixref{section-appendix-training-process}, \algoref{alg:training_process} shows the training process of CTTSP.

\section{Experiments}
\label{section-5}

\subsection{Descriptions of Datasets}
We conduct experiments on four benchmarks:
\begin{itemize}
    \item  \textbf{JingDong}\footnote{https://jdata.jd.com/html/detail.html?id=8} contains the actions of users about purchasing, browsing, commenting, following and adding products to shopping carts. We select the purchasing actions in March 2018 and treat products bought on the same day by each user as a set.
   
    \item  \textbf{Dunnhumby-Carbo (DC)}\footnote{https://www.dunnhumby.com/careers/engineering/sourcefiles} records the transactions of households at a retailer in two years. We select transactions in the first 60 days to conduct experiments and treat products purchased on the same day by each household as a set. 
       
	\item \textbf{TaFeng}\footnote{https://www.kaggle.com/chiranjivdas09/ta-feng-grocery-dataset} includes the shopping transactions at a Chinese grocery store from November 2000 to February 2001. Products bought on the same day by each customer are treated as a set.
    
    \item  \textbf{TaoBao}\footnote{https://tianchi.aliyun.com/dataset/dataDetail?dataId=649} records the online user behaviors about purchasing, clicking, marking products as favors, and adding products to shopping carts. We select the purchasing behaviors and treat categories of products purchased on the same day by each user as a set. This dataset is recorded from November 24, 2017 to December 3, 2017.
\end{itemize}
We strictly follow \citet{DBLP:conf/www/YuWS0L22} to preprocess the datasets. We select the frequent elements that cover 80\% records in each dataset. We drop the sequences with lengths less than 4 and crop the sequences with lengths more than 20. Note that the TMS dataset in\citet{DBLP:conf/www/YuWS0L22} is not used because the absolute temporal information is unavailable, but we additionally use the TaFeng dataset to do experiments.
We compare our method with the existing methods under both transductive and inductive settings. For the transductive setting, we follow \citet{DBLP:conf/www/YuWS0L22} to use the last set, the second last set, and the remaining sets of each user for testing, validation, and training. 
For the inductive setting, we follow \citet{DBLP:conf/kdd/YuSDL0L20} to randomly divide each dataset into the training, validation, and testing sets with the ratio of 70\%, 10\%, and 20\% across users. 
We provide statistics of the datasets in \appendixref{section-appendix-data-statistics}.

\subsection{Compared Baselines}
Our approach is compared with the following baselines:
\begin{itemize}
    \item  \textbf{TOP} uses the most frequent elements in all the users' sequences as the predictions for any user.

    \item  \textbf{PTOP} predicts the most frequent elements in the individual sequence for each user.

    \item  \textbf{FPMC} utilizes the matrix factorization and Markov chain to predict the next-period basket, which can capture the element transitions as well as long-term user preferences \cite{DBLP:conf/www/RendleFS10}.

    \item  \textbf{DREAM} first uses pooling operations to embed baskets into vectors, and then feeds the sequence of basket embeddings into an RNN to predict the next-period basket \cite{DBLP:conf/sigir/YuLWWT16}.

    \item  \textbf{TGN} is a general dynamic graph learning framework, which computes evolving representations for nodes in the graph \cite{DBLP:journals/corr/abs-2006-10637}. To adapt TGN for the temporal sets prediction task, we treat each user-set interaction as multiple user-element interactions and separately learn from the user-element interactions. 

    \item  \textbf{RUM} is a memory-augmented neural network for sequential recommendation, aiming to exploit sequential behaviors of users \citet{DBLP:conf/wsdm/ChenXZT0QZ18}. We first use the pooling operation to embed each set and then employ RUM to handle the sequences of set embeddings.
    
    \item  \textbf{HPMN} introduces a hierarchical framework with an updating mechanism for multiple periods, which can capture the multi-scale sequential patterns in the lifelong behavior sequences of users \citet{DBLP:conf/sigir/RenQF0ZBZXYZG19}. We incorporate the pooing operations into HPMN to apply it for predicting temporal sets.
    
    \item  \textbf{DIN} is designed for the click-through rate prediction, which learns the representation of user interests with respect to each ad by the attention mechanism \cite{DBLP:conf/kdd/ZhouZSFZMYJLG18}. We adapt DIN by splitting each set into separate elements and learning from the sequences of elements.

    \item  \textbf{Sets2Sets} first generates the representations of sets via the average pooling and then captures temporal dependencies with an encoder-decoder framework for predicting multi-period sets. It also considers the repeated patterns in user behaviors \cite{DBLP:conf/kdd/HuH19}.

    \item  \textbf{DSNTSP} designs a dual transformer-based architecture to learn both element-level and set-level representations for each user's sequence. A co-transformer module is further proposed to capture the multiple temporal dependencies of elements and sets \cite{DBLP:conf/sigir/SunBDL0L20}.
    
    \item  \textbf{DNNTSP} constructs set-level co-occurrence graphs to learn the element relationships and uses the attention mechanism to capture temporal dependencies. The gating mechanism is also employed to discover the shared patterns among elements \cite{DBLP:conf/kdd/YuSDL0L20}.
    
    \item  \textbf{ETGNN} first connects the sequences of different users via a temporal graph and then learns from the graph according to an element-guided message aggregation mechanism and a temporal information utilization component \cite{DBLP:conf/www/YuWS0L22}. 
\end{itemize}

\begin{table*}[!htbp]
\centering
\caption{Performance of different methods on all the datasets. The best and second-best performance are boldfaced and underlined. "Improv." denotes the improvement over the best baseline, and ${\star}$ indicates the improvement is statistically significant with a paired t-test (i.e., $p$ \textless 0.05).}
\label{tab:performance_comparison}
\resizebox{\textwidth}{!}
{
\setlength{\tabcolsep}{0.2mm}
{
\begin{tabular}{c|c|ccc|ccc|ccc|ccc}
\hline
\multirow{2}{*}{Datasets}  & \multirow{2}{*}{Methods} & \multicolumn{3}{c|}{$K$=10} & \multicolumn{3}{c|}{$K$=20} & \multicolumn{3}{c|}{$K$=30} & \multicolumn{3}{c}{$K$=40} \\ \cline{3-14} 
                           &                          & Recall          & NDCG             & PHR             & Recall          & NDCG            & PHR             & Recall          & NDCG            & PHR             & Recall          & NDCG            & PHR  \\ \hline
\multirow{14}{*}{JingDong} & TOP                      &  {0.1531}         &  {0.0988}       &  {0.1574}      &     {0.1826}      &  {0.1076}       &  {0.1926}      &  {0.2115}         &  {0.1143}       &  {0.2207}      &  {0.2395}         &     {0.1198}    &     {0.2484}  \\
                           & PTOP              &  {0.2709}         &  {0.2264}       &  {0.2905}      &     {0.2742}      &  {0.2276}       &  {0.2935}      &  {0.2757}         &  {0.2279}       &  {0.2954}      &  {0.2762}         &     {0.2280}    &     {0.2964}  \\
                           & FPMC                     &  {0.2704}         &  {0.2109}       &  {0.2880}      &     {0.2973}      &  {0.2182}       &  {0.3134}      &  {0.3082}         &  {0.2207}       &  {0.3245}      &  {0.3178}         &     {0.2226}    &     {0.3346}  \\
                           & DREAM                    &  {0.2888}         &  {0.2198}       &  {0.3033}      &     {0.3373}      &  {0.2329}       &  {0.3513}      &  {0.3637}         &  {0.2388}       &  {0.3787}      &  {0.3757}         &     {0.2413}    &     {0.3918}  \\
                           & TGN                      &  {0.2359}         &  {0.1682}       &  {0.2491}      &     {0.2817}      &  {0.1807}       &  {0.2977}      &  {0.3090}         &  {0.1870}       &  {0.3262}      &  {0.3280}         &     {0.1909}    &     {0.3461}  \\
                           & RUM                      &  {0.2921}         &  {0.2229}       &  {0.3105}      &     {0.3269}      &  {0.2326}       &  {0.3448}      &  {0.3429}         &  {0.2363}       &  {0.3608}      &  {0.3560}         &     {0.2390}    &     {0.3735}  \\
                           & HPMN                      &  {0.2582}         &  {0.2131}       &  {0.2762}      &     {0.2754}      &  {0.2179}       &  {0.2948}      &  {0.2824}         &  {0.2196}       &  {0.3026}      &  {0.2913}         &     {0.2214}    &     {0.3115}  \\
                           & DIN                      &  {0.3024}         &  {0.2503}       &  {0.3213}      &     {0.3176}      &  {0.2545}       &  {0.3379}      &  {0.3262}         &  {0.2565}       &  {0.3461}      &  {0.3328}         &     {0.2580}    &     {0.3519}  \\
                           & Sets2Sets                &  {0.3209}         &  {0.2497}       &  {0.3418}      &     {0.3474}      &  {0.2571}       &  {0.3696}      &  {0.3623}         &  {0.2604}       &  {0.3843}      &  {0.3735}         &     {0.2627}    &     {0.3960}  \\
                           & DSNTSP                   & {0.3464} & \underline{0.2734} & {0.3670} & {0.3750} & {0.2820} & {0.3947} & {0.3883} & {0.2852} & {0.4078} & {0.3963} & {0.2869} & {0.4150}  \\
                           & DNNTSP                   & {0.3224}         &  {0.2458}       &  {0.3470}      &     {0.3568}      &  {0.2551}       &  {0.3813}      &  {0.3747}         &  {0.2594}       &  {0.3986}      &  {0.3843}         &     {0.2613}    &     {0.4074}  \\
                           & ETGNN                  &      \underline{0.3658}     &   {0.2724}     &   \underline{0.3885}     &     \underline{0.4217}      &     \underline{0.2878}   &     \underline{0.4460}   &   \underline{0.4558}        &      \underline{0.2956}   &     \underline{0.4780}   &     \underline{0.4752}      &    \underline{0.2997}     &   \underline{0.4959} \\
                           & CTTSP                   & \textbf{0.4078}$^{\star}$ & \textbf{0.2946}$^{\star}$ & \textbf{0.4311}$^{\star}$ & \textbf{0.4507}$^{\star}$ & \textbf{0.3072}$^{\star}$ & \textbf{0.4724}$^{\star}$ & \textbf{0.4704}$^{\star}$ & \textbf{0.3118}$^{\star}$ & \textbf{0.4911}$^{\star}$ & \textbf{0.4895}$^{\star}$ & \textbf{0.3155}$^{\star}$ & \textbf{0.5105}$^{\star}$   \\ 
                           & Improv.              &  {11.48\%}         &  {7.75\%}       &  {10.97\%}      &     {6.88\%}      &  {6.74\%}       &  {5.92\%}      &  {3.20\%}         &  {5.48\%}       &  {2.74\%}      &  {3.01\%}         &     {5.27\%}    &     {2.94\%}  \\ \hline
\multirow{14}{*}{DC}       & TOP                      & {0.1606}        & {0.0839}         & {0.2326}      &     {0.2521}      &  {0.1093}       &  {0.3430}      &  {0.3279}         &  {0.1269}       &  {0.4251}      &  {0.3872}         &     {0.1397}    &     {0.4906}  \\
                           & PTOP              & {0.4080}        & {0.3161}         & {0.5039}      &     {0.4383}      &  {0.3246}       &  {0.5389}      &  {0.4636}         &  {0.3306}       &  {0.5663}      &  {0.4982}         &     {0.3381}    &     {0.5980}  \\
                           & FPMC                     & {0.2462}        & {0.1991}         & {0.3274}      &     {0.3175}      &  {0.2191}       &  {0.4128}      &  {0.3771}         &  {0.2332}       &  {0.4805}      &  {0.4323}         &     {0.2451}    &     {0.5386}  \\
                           & DREAM                    & {0.3159}        & {0.2266}         & {0.4091}      &     {0.4102}      &  {0.2532}       &  {0.5089}      &  {0.4813}         &  {0.2701}       &  {0.5821}      &  {0.5427}         &     {0.2833}    &     {0.6428}  \\
                           & TGN                      & {0.3230}        & {0.2359}         & {0.4141}      &     {0.4232}      &  {0.2639}       &  {0.5267}      &  {0.4914}         &  {0.2798}       &  {0.5941}      &  {0.5445}         &     {0.2913}    &     {0.6453}  \\
                           & RUM                      &  {0.3281}         &  {0.2551}       &  {0.4184}      &     {0.3955}      &  {0.2740}       &  {0.4926}      &  {0.4486}         &  {0.2866}       &  {0.5505}      &  {0.4967}         &     {0.2970}    &     {0.6008}  \\
                           & HPMN                      &  {0.2560}         &  {0.1962}       &  {0.3403}      &     {0.3373}      &  {0.2188}       &  {0.4357}      &  {0.4005}         &  {0.2338}       &  {0.5044}      &  {0.4499}         &     {0.2444}    &     {0.5575} \\
                           & DIN                      & {0.3747}        & {0.2761}         & {0.4752}      &     {0.4617}      &  {0.3004}       &  {0.5673}      &  {0.5166}         &  {0.3135}       &  {0.6222}      &  {0.5687}         &     {0.3247}    &     {0.6717}  \\
                           & Sets2Sets                & {0.4417}        & {0.3169}         & {0.5383}      &     {0.5031}      &  {0.3342}       &  {0.6004}      &  {0.5533}         &  {0.3459}       &  {0.6514}      &  {0.5936}         &     {0.3546}    &     {0.6913}  \\
                           & DSNTSP                   & {0.4399}        & {0.3201}       &  {0.5386}      &     {0.5112}      &  {0.3303}       &  {0.6105}      &  {0.5615}         &  {0.3522}       &  {0.6608}      &  {0.6031}         &     {0.3612}    &     {0.7004}  \\
                           & DNNTSP                   & {0.4461} &  {0.3176} &  {0.5442} & {0.5168} & {0.3374} & {0.6170} & {0.5634} & {0.3483} & {0.6626} &  {0.6067}         &     {0.3575}    &     {0.7033}  \\
                           & ETGNN                  &   \underline{0.4593}        &   \underline{0.3321}    &    \underline{0.5582}    &    \underline{0.5477}       &    \underline{0.3567}     &    \underline{0.6454}    &     \underline{0.6070}      &    \underline{0.3708}     &     \underline{0.7009}   &      \underline{0.6580}     &    \underline{0.3818}     &    \underline{0.7468} \\
                           & CTTSP                   & \textbf{0.4672}$^{\star}$ & \textbf{0.3337}$^{\star}$ & \textbf{0.5669}$^{\star}$ & \textbf{0.5576}$^{\star}$ & \textbf{0.3592}$^{\star}$ & \textbf{0.6546}$^{\star}$ & \textbf{0.6209}$^{\star}$ & \textbf{0.3740}$^{\star}$     &     \textbf{0.7128}$^{\star}$   &      \textbf{0.6691}$^{\star}$     &    \textbf{0.3844}$^{\star}$     &    \textbf{0.7545}$^{\star}$   \\ 
                           & Improv.              &  {1.72\%}         &  {0.48\%}       &  {1.56\%}      &     {1.81\%}      &  {0.70\%}       &  {1.43\%}      &  {2.29\%}         &  {0.86\%}       &  {1.70\%}      &  {1.69\%}         &     {0.68\%}    &     {1.03\%}  \\ \hline                             
\multirow{14}{*}{TaFeng}   & TOP                      & {0.0896}        & {0.0965}         & {0.2644}        & {0.1207}        & {0.1050}        & {0.3463}        & {0.1469}        & {0.1134}        & {0.4105}        & {0.1586}        & {0.1171}        & {0.4391} \\
                           & PTOP              & {0.1282}        & {0.1158}         & {0.3770}        & {0.1721}        & {0.1293}        & {0.4685}        & {0.1959}        & {0.1364}        & {0.5110}        & {0.2105}        & {0.1413}        & {0.5336} \\
                           & FPMC                     & {0.0675}        & {0.0553}         & {0.2036}        & {0.0932}        & {0.0633}        & {0.2766}        & {0.1121}        & {0.0692}        & {0.3266}        & {0.1274}        & {0.0738}        & {0.3644} \\
                           & DREAM                    & {0.0977}        & {0.0928}         & {0.2760}        & {0.1214}        & {0.1001}        & {0.3504}        & {0.1404}        & {0.1063}        & {0.4011}        & {0.1585}        & {0.1119}        & {0.4462} \\
                           & TGN                      & {0.0987}        & {0.0909}         & {0.2989}        & {0.1251}        & {0.0985}        & {0.3669}        & {0.1414}        & {0.1037}        & {0.4061}        & {0.1558}        & {0.1078}        & {0.4382}\\        
                           & RUM                      &  {0.1115}         &  {0.0930}       &  {0.3317}      &     {0.1473}      &  {0.1037}       &  {0.4121}      &  {0.1699}         &  {0.1107}       &  {0.4564}      &  {0.1855}         &     {0.1154}    &     {0.4895}  \\
                           & HPMN                      &  {0.0932}         &  {0.0800}       &  {0.2711}      &     {0.1188}      &  {0.0878}       &  {0.3454}      &  {0.1389}         &  {0.0942}       &  {0.4000}      &  {0.1561}         &     {0.0994}    &     {0.4383} \\                   
                           & DIN                      & {0.0997}        & {0.0826}         & {0.3005}        & {0.1276}        & {0.0907}        & {0.3751}        & {0.1456}        & {0.0963}        & {0.4204}        & {0.1610}        & {0.1009}        & {0.4551} \\
                           & Sets2Sets                & \underline{0.1555}        & \underline{0.1309}         & \underline{0.4327}        & \underline{0.2149}        & \underline{0.1499}        & \underline{0.5466}        & \underline{0.2486}        & \underline{0.1609}        & \underline{0.6014}        & \underline{0.2721}        & \underline{0.1683}        & \underline{0.6367} \\
                           & DSNTSP                   & {0.1404}        & {0.1153}         & {0.4043}        & {0.1857}        & {0.1291}        & {0.4991}        & {0.2126}        & {0.1376}        & {0.5488}        & {0.2329}        & {0.1437}        & {0.5843} \\
                           & DNNTSP                   & {0.1475}        & {0.1160}         & {0.4127}        & {0.2056}        & {0.1350}        & {0.5311}        & {0.2359}        & {0.1452}        & {0.5864}        & {0.2579}        & {0.1522}        & {0.6198} \\
                           & ETGNN                 &  {0.1450}         &  {0.1123}       &  {0.4151}      &     {0.2040}      &  {0.1319}       &  {0.5327}      &  {0.2395}         &  {0.1434}       &  {0.5916}      &  {0.2627}         &     {0.1505}    &     {0.6235}  \\
                           & CTTSP                   & \textbf{0.1618}$^{\star}$ & \textbf{0.1353}$^{\star}$  & \textbf{0.4428}$^{\star}$ & \textbf{0.2249}$^{\star}$ & \textbf{0.1563}$^{\star}$ & \textbf{0.5582}$^{\star}$ & \textbf{0.2603}$^{\star}$ & \textbf{0.1682}$^{\star}$ & \textbf{0.6140}$^{\star}$ & \textbf{0.2883}$^{\star}$ & \textbf{0.1769}$^{\star}$ & \textbf{0.6506}$^{\star}$ \\
                           & Improv.              & {4.05\%}              & {3.36\%}               & {2.33\%}              & {4.65\%}              & {4.27\%}              & {2.12\%}              & {4.71\%}              & {4.54\%}              & {2.10\%}              & {5.95\%}              & {5.11\%}              & {2.18\%} \\ \hline
\multirow{14}{*}{TaoBao}   & TOP                      &  {0.1572}         &  {0.0835}       &  {0.1987}      &     {0.2457}      &  {0.1074}       &  {0.2964}      &  {0.3091}         &  {0.1220}       &  {0.3637}      &  {0.3609}         &     {0.1328}    &     {0.4208}  \\
                           & PTOP              &  {0.1794}         &  {0.1240}       &  {0.2187}      &     {0.1909}      &  {0.1272}       &  {0.2328}      &  {0.1984}         &  {0.1289}       &  {0.2424}      &  {0.2061}         &     {0.1305}    &     {0.2517}  \\
                           & FPMC                     &  {0.1675}         &  {0.0959}       &  {0.2088}      &     {0.2548}      &  {0.1196}       &  {0.3082}      &  {0.3189}         &  {0.1343}       &  {0.3778}      &  {0.3659}         &     {0.1440}    &     {0.4273}  \\
                           & DREAM                    &  {0.1665}         &  {0.0932}       &  {0.2069}      &     {0.2566}      &  {0.1177}       &  {0.3079}      &  {0.3185}         &  {0.1319}       &  {0.3752}      &  {0.3663}         &     {0.1419}    &     {0.4262}  \\
                           & TGN                      &  {0.1665}         &  {0.0888}       &  {0.2082}      &     {0.2560}      &  {0.1129}       &  {0.3088}      &  {0.3226}         &  {0.1283}       &  {0.3799}      &  {0.3727}         &     {0.1387}    &     {0.4329}  \\
                           & RUM                      &  {0.1671}         &  {0.1077}       &  {0.2072}      &     {0.2301}      &  {0.1248}       &  {0.2797}      &  {0.2734}         &  {0.1347}       &  {0.3280}      &  {0.3096}         &     {0.1423}    &     {0.3678}  \\
                           & HPMN                      &  {0.1848}         &  {0.1144}       &  {0.2281}      &     {0.2692}      &  {0.1374}       &  {0.3235}      &  {0.3310}         &  {0.1516}       &  {0.3900}      &  {0.3775}         &     {0.1612}    &     {0.4394} \\
                           & DIN                      &  {0.2188}         &  {0.1317}       &  {0.2671}      &     {0.3056}      &  {0.1553}       &  {0.3623}      &  {0.3646}         &  {0.1690}       &  {0.4255}      &  {0.4088}         &     {0.1782}    &     {0.4716}  \\
                           & Sets2Sets                &  {0.2413}         &  {0.1488}       &  {0.2911}      &     {0.3228}      &  {0.1710}       &  {0.3821}      &  {0.3838}         &  {0.1850}       &  {0.4465}      &  {0.4315}         &     {0.1950}    &     {0.4954}  \\
                           & DSNTSP                   &  {0.2363}         &  {0.1431}       &  {0.2867}      &     {0.3296}      &  {0.1685}       &  {0.3885}      &  {0.3932}    &  {0.1832}       &  {0.4557}      &  {0.4414}         &     {0.1932}    &     {0.5050}  \\
                          & DNNTSP                   & {0.2511} &  {0.1535}     &  {0.3028}     &     {0.3369}    &  {0.1769}     &  {0.3972}   &  {0.3925}  &  {0.1898} &  {0.4535} &  {0.4384} &     {0.1994}    &     {0.5024}  \\
                          & ETGNN                  &     \underline{0.2589}      &    \underline{0.1542}     &     \underline{0.3103}   &     \underline{0.3525}      &    \underline{0.1798}     &    \underline{0.4134}    &     \underline{0.4124}      &    \underline{0.1937}     &    \underline{0.4760}    &      \underline{0.4596}     &    \underline{0.2036}     &    \underline{0.5239} \\                           
                          & CTTSP                   & \textbf{0.2610}$^{\star}$ &   \textbf{0.1595}$^{\star}$  &  \textbf{0.3134}$^{\star}$  &  \textbf{0.3558}$^{\star}$ &  \textbf{0.1855}$^{\star}$  &  \textbf{0.4167}$^{\star}$  & \textbf{0.4170}$^{\star}$  &  \textbf{0.1996}$^{\star}$  &   \textbf{0.4801}$^{\star}$   &      \textbf{0.4627}$^{\star}$     &    \textbf{0.2092}$^{\star}$     &    \textbf{0.5264}$^{\star}$   \\ 
                          & Improv.              &  {0.81\%}         &  {3.44\%}       &  {1.00\%}      &     {0.94\%}      &  {3.17\%}       &  {0.80\%}      &  {1.12\%}         &  {3.05\%}       &  {0.86\%}      &  {0.67\%}         &     {2.75\%}    &     {0.48\%}  \\ \hline
\end{tabular}
}
}
\end{table*}

\subsection{Experimental Settings}
\label{sec:experimental_settings}
To evaluate the model performance, we follow \citet{DBLP:conf/kdd/YuSDL0L20,DBLP:conf/www/YuWS0L22} to rank the top-$K$ elements using the predicted probabilities and set $K$ to 10, 20, 30, and 40. We adopt Recall, Normalized Discounted Cumulative Gain (NDCG), and Personal Hit Ratio (PHR) as the evaluation metrics. 
Adam optimizer \cite{DBLP:journals/corr/KingmaB14} is leveraged to optimize the models. We employ the cosine annealing learning rate scheduler \cite{DBLP:conf/iclr/LoshchilovH17} to change the learning rate.
Dropout \cite{DBLP:journals/jmlr/SrivastavaHKSS14} is adopted to prevent models from over-fitting. We train the methods with a fixed 2000 epochs and use an early stopping strategy with a patience of 100. The model with the best performance (more specifically, the highest average NDCG) on the validation set is used for testing. To eliminate the deviations, we run the methods ten times with seeds from 0 to 9 on all the datasets and report the average performance. For the proposed CTTSP, we set the learning rate to 0.001 on all the datasets. We apply the grid search to find the best settings of CTTSP. Specifically, the dropout rate and hidden dimension $d$ are searched in [0.0, 0.05, 0.1, 0.15, 0.2] and [32, 64, 128]. The hyperparameter $\lambda_{up}$ and $\lambda_{cp}$ are both searched in [0.0, 0.05, 0.1, 0.3, 0.5, 0.7, 0.9, 0.95, 1.0]. The settings of our model on different datasets are shown in \appendixref{section-appendix-hyper-parameters}. 

Note that the memories of users and elements are continuously updated during the training, validation, and testing process, but the model parameters are not optimized in the validation and testing process to avoid the data leakage issue.
We train our model in a mini-batch manner, where the batches are created by our set-batch algorithm in advance. All the experiments are conducted on an Ubuntu machine equipped with one Intel(R) Xeon(R) Gold 6130 CPU @ 2.10GHz, which has 16 physical cores. The GPU device is NVIDIA Tesla T4 with 15 GB memory.  Our approach is implemented with PyTorch \cite{DBLP:conf/nips/PaszkeGMLBCKLGA19}.
Codes and datasets are available at https://github.com/yule-BUAA/CTTSP.

\subsection{Performance under Transductive Setting}
The comparisons of our approach with baselines under the transductive setting are shown in \tabref{tab:performance_comparison}. From \tabref{tab:performance_comparison}, several conclusions can be summarized as follows.

Firstly, PTOP usually shows better performance than TOP because PTOP recommends personalized elements for each user based on his/her individual sequence, while TOP just identically predicts the most frequent elements for all the users which leads to unsatisfactory results. This reveals the importance of learning each user's personal preference for temporal sets prediction.

Secondly, DREAM is often superior to FPMC because FPMC only utilizes the Markov chain to capture adjacent behaviors while DREAM employs the RNNs to capture temporal dependencies in the whole sequence of each user. This indicates that it is necessary to comprehensively mine the temporal information in users' historical behaviors.

Thirdly, TGN, RUM, HPMN, and DIN often achieve competitive or better results than the above baselines, indicating some designs for memory networks or recommender systems can also facilitate the prediction of temporal sets. RUM and HPMN perform better than TGN in most cases as they design specialized memory updating mechanisms (i.e., item-level and feature-level versions in RUM, hierarchical and periodical mechanisms in HPMN). DIN learns element-specific representations for each user, which can effectively enhance the expressive ability of the model. However, all of them fail to handle the unique properties of sets and there is still space for improvement.

Fourthly, Sets2Sets, DSNTSP, DNNTSP, and ETGNN, which are customized approaches for the temporal sets prediction problem, usually yield better performance than other baselines. In particular, Sets2Sets additionally explores repeated patterns in user behaviors. DSNTSP learns both the element-level and set-level representations for the sequence of each user. DNNTSP studies the correlations of elements within each set and captures the temporal dependencies of elements across different sets. ETGNN is the best baseline in most cases because it not only learns element-specific representations that are aware of the temporal information for each user but also captures the collaborative signals in high-order user-element interactions. 

Finally, CTTSP significantly outperforms the existing methods over all the metrics on all the datasets. The superiority of our approach lies in 1) the continuous-time learning framework to explicitly model the evolving user preferences and exploit the latent collaborative signals among different users; 2) the learning of personalized user behaviors, which can aggregate the previously interacted elements of each user according to the user and elements.

\subsection{Performance under Inductive Setting}
\begin{figure}[!htbp]
    \centering
    \includegraphics[width=0.98\columnwidth]{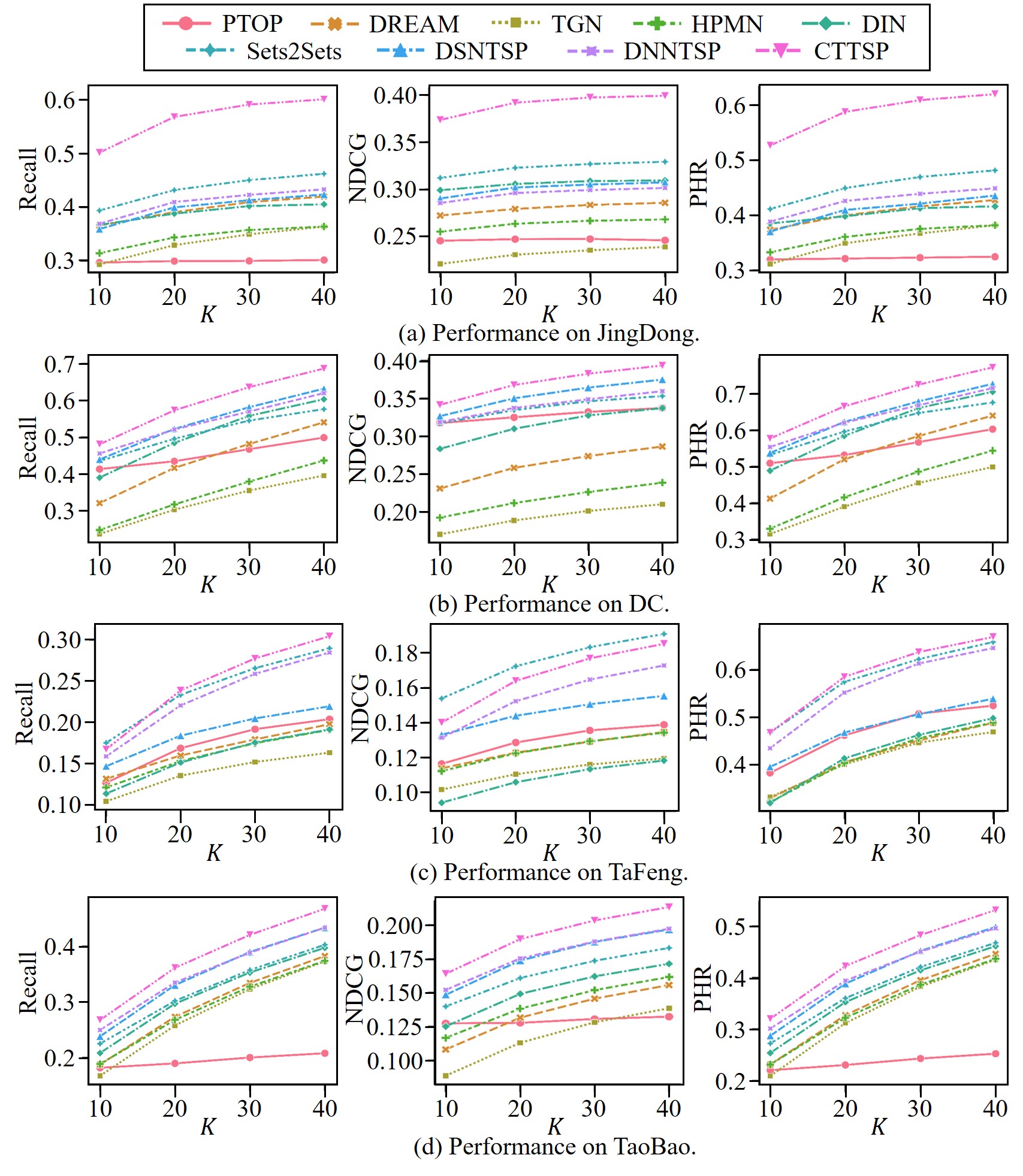}
    \caption{Performance of different methods under the inductive setting. The performance of TOP is not depicted due to its inferior performance.}
    \label{fig:inductive_results}
\end{figure}

CTTSP is endowed with the inductive ability because 1) the memory of each user in the memory bank is identically initialized as a zero vector at first. When a new user comes, his/her memory can be updated according to the learned model parameters; 2) the static embedding $\bm{e}^u$ is shared across the users in \secref{sec:user_pespective_sec}, which is not influenced by new users. 
Therefore, we evaluate the model performance under the inductive setting, which predicts the next-period set for users \textit{that are not contained in the training set}. Note that FPMC, RUM, and ETGNN could only be evaluated under the transductive setting because they contain user-specific trainable parameters and are inherently not inductive. We report the performance of different methods in \figref{fig:inductive_results}.

From \figref{fig:inductive_results}, we find that the results of baselines vary across different datasets, but our CTTSP often consistently achieves better performance than baselines. The superiority of our approach under the inductive setting may be due to the learning of memories of users and elements, which can help our model discover the correlations across different users, and thus increase the reasoning ability for new users. This advantage indicates the potential of CTTSP in tackling the cold-start problems \cite{DBLP:conf/kdd/LeeIJCC19}. 
We also observe that Sets2Sets usually performs better than baselines and it even achieves higher NDCG than our approach on TaFeng. We owe such a phenomenon to the repeated user behavior learning module in Sets2Sets, which is similar to the PerTOP baseline but with trainable importance weights. PerTOP achieves relatively good NDCG on TaFeng, and its trainable version in Sets2Sets may contribute more and lead to higher NDCG.

\subsection{Effects of Continuous-Time and Personalized Information of Users}
\label{sec:evolutionary_stationary_preferences}
\begin{figure}[!htbp]
    \centering
    \includegraphics[width=1.0\columnwidth]{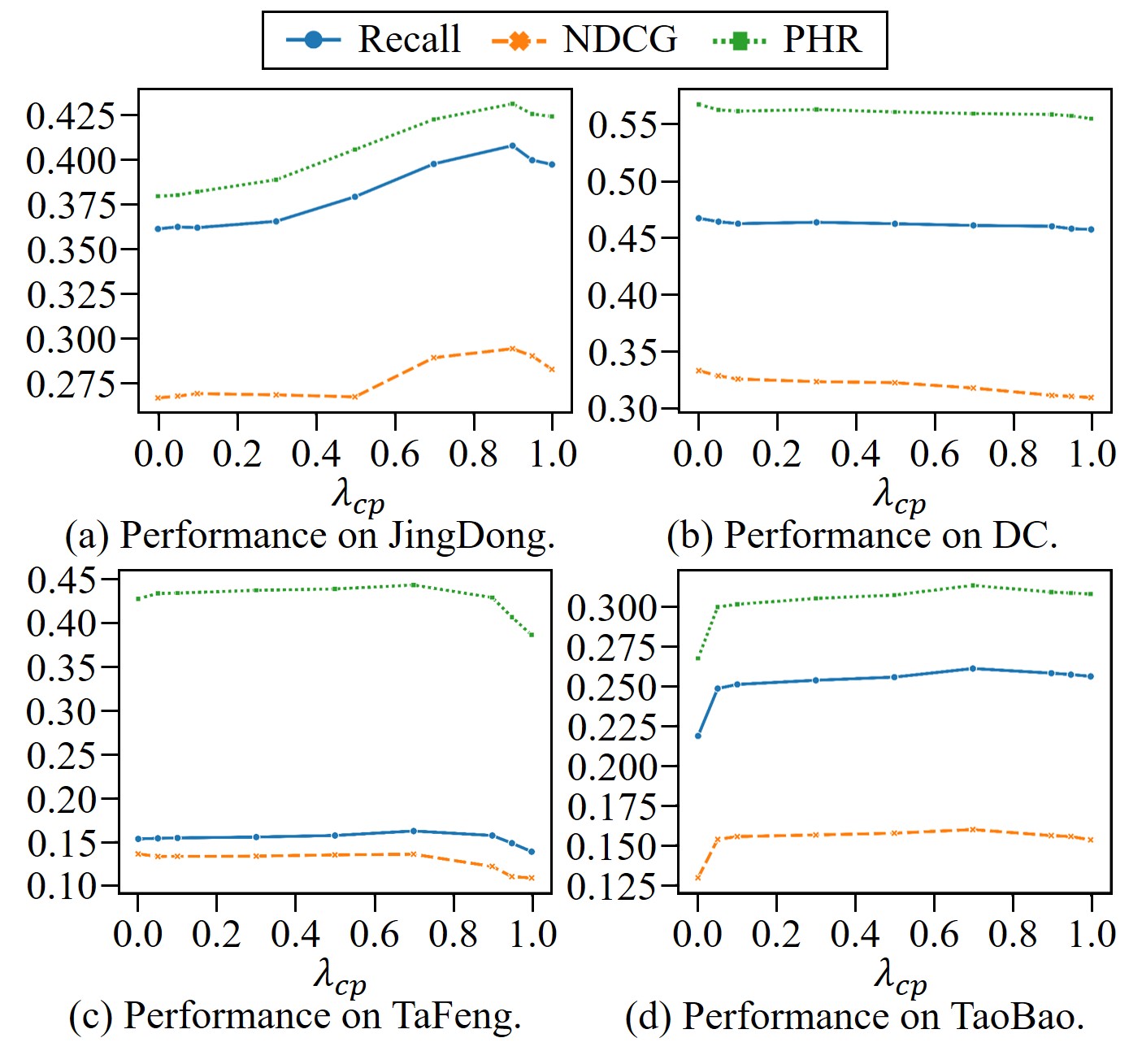}
    \caption{Effects of the continuous-time and personalized information of users on different datasets.}
    \label{fig:evolutionary_preference_importance}
\end{figure}

We further investigate the importance of continuous-time and personalized information of users. In particular, we vary the continuous-time probability importance $\lambda_{cp}$ in [0.0, 0.05, 0.1, 0.3, 0.5, 0.7, 0.9, 0.95, 1.0] by fixing the user perspective importance $\lambda_{up}$ as its best configuration. Due to space limitations, we report the results on all the datasets in \figref{fig:evolutionary_preference_importance}, where $K$ is set to 10. Similar trends could be observed when the values of $K$ are set to 20, 30, and 40.

From \figref{fig:evolutionary_preference_importance}, we can observe that the best settings of $\lambda_{cp}$ change over different datasets and reflect the datasets' unique properties. Two representative datasets are JingDong and DC, where the continuous-time user preferences are more obvious on JingDong, while DC is more likely to be affected by personalized user behaviors. TaFeng and TaoBao tend to show the balance of both continuous-time and personalized information of users. These findings suggest that it is necessary to select appropriate $\lambda_{cp}$ according to the property of each dataset.

\subsection{Effects of User and Element Perspectives}
\label{sec:user_element_perspectives}
\begin{figure}[!htbp]
    \centering
    \includegraphics[width=1.0\columnwidth]{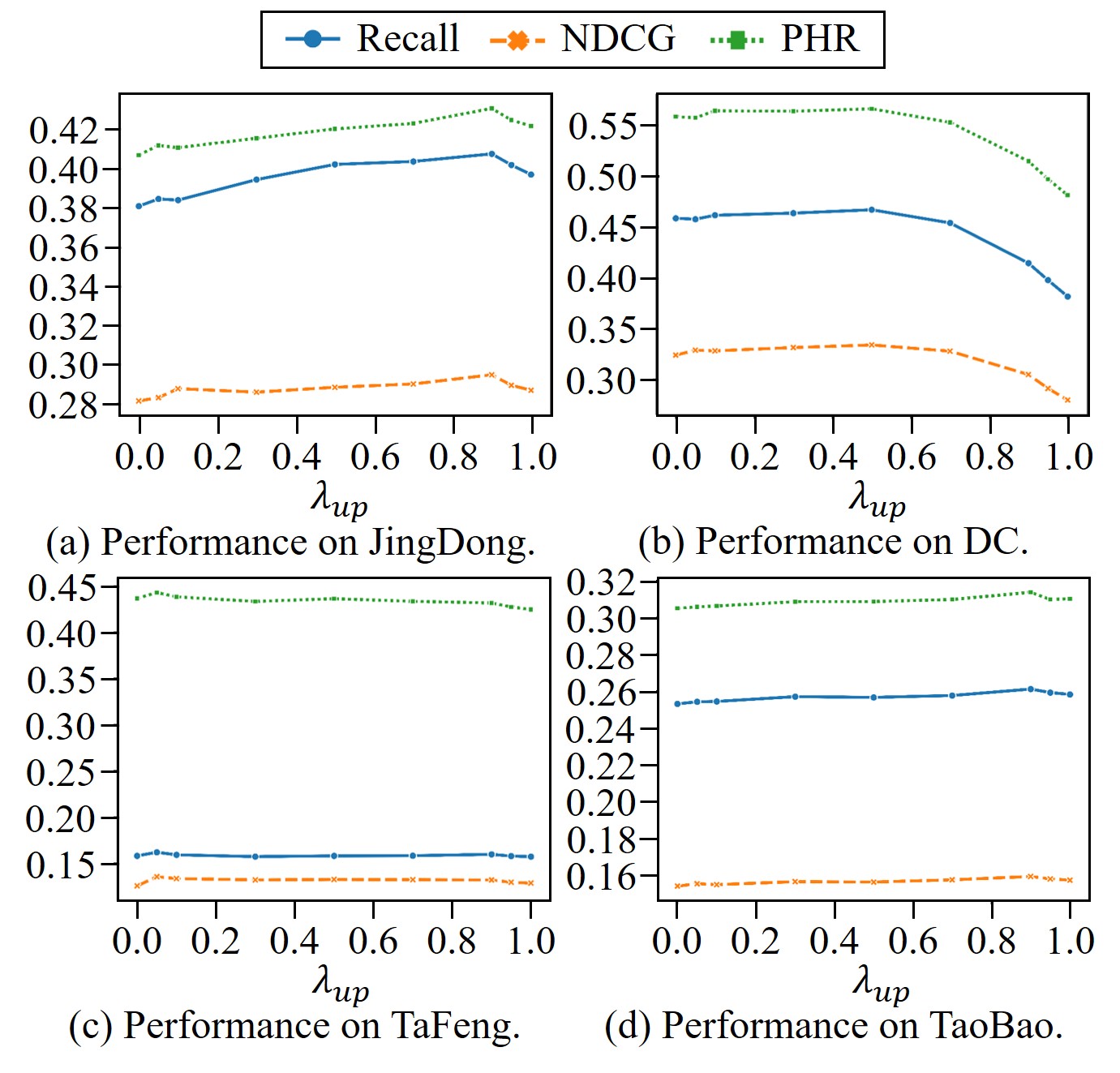}
    \caption{Effects of user and element perspectives on different datasets.}
    \label{fig:user_perspective_importance}
\end{figure}

We also study the impacts of the user and element perspectives on different datasets. Specifically, we first fix the continuous-time probability importance $\lambda_{cp}$ as its best configuration and then change the value of the user perspective importance $\lambda_{up}$ in [0.0, 0.05, 0.1, 0.3, 0.5, 0.7, 0.9, 0.95, 1.0]. Experimental results on all the datasets are depicted in \figref{fig:user_perspective_importance}, where $K$ is set to 10.

From \figref{fig:user_perspective_importance}, we conclude that different datasets depict varied importance of the user perspective and element perspective. Specifically, JingDong and TaoBao are more likely to be influenced by the user perspective, while TaFeng pays more attention to the element perspective. DC keeps a balance of the user and element perspective. Therefore, choosing suitable $\lambda_{up}$ on different datasets is also essential for satisfactory prediction performance.

\subsection{Ablation Study}
\begin{figure}[!htbp]
    \centering
    \includegraphics[width=1.0\columnwidth]{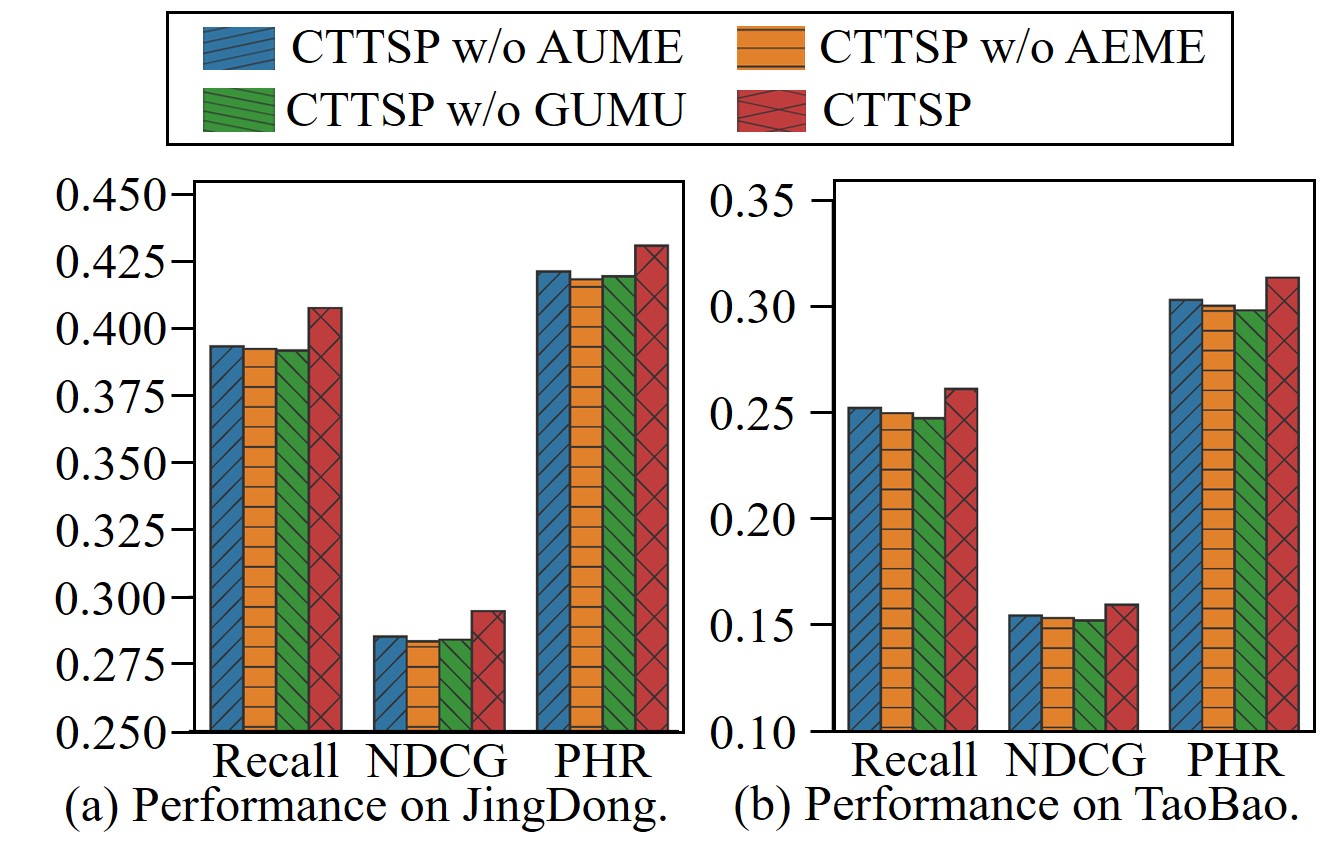}
    \caption{Ablation study of the proposed model on JingDong and TaoBao.}
    \label{fig:ablation_study}
\end{figure}

We further conduct the ablation study to validate the effectiveness of the Attention mechanism in the User Message Encoder (AUME), the Attention mechanism in the Element Message Encoder (AEME), and the Gated Updating mechanism in the Memory Updaters (GUMU). 
Specifically, we use the average pooling operation over the related elements in each user-set interaction when computing the user's message and denote the model as CTTSP w/o AUME. This indicates that CTTSP w/o AUME does not distinguish the importance of different elements to the user. CTTSP w/o AEME is implemented by encoding the message of each element according to the information of the interacted user and the element itself, while the information of other elements in the same user-set interaction is ignored. We employ Gated Recurrent Units (GRU) in \citet{DBLP:conf/emnlp/ChoMGBBSB14} to replace the GUMU component for both users and elements, and denote the model as CTTSP w/o GUMU. Experimental results of the variants on JingDong and TaoBao are shown in \figref{fig:ablation_study}, where $K$ is set to 10.

From \figref{fig:ablation_study}, we can observe that CTTSP achieves the best performance when it uses all the three components, and removing any component would lead to worse results. In particular, the AUME component encodes the user's message by learning the importance of different elements. The AEME component captures the intra-set relationships of elements when encoding the message of each element. The GUMU component adaptively selects the encoded message and historical memories according to the gated updating mechanism with fewer parameters than GRU, which may reduce the training difficulty \cite{DBLP:conf/icml/PascanuMB13}.

\subsection{Efficiency of the Set-batch Algorithm}
We validate the efficiency of the proposed set-batch algorithm by comparing the running time of our CTTSP with CTTSP w/o set-batch. It is worth noticing that the set-batch algorithm can create time-consistent batches during the preprocessing phase in advance, and thus does not introduce extra cost in the training and evaluation process. Moreover, there is no need to predetermine the number of batches as it is automatically provided by the set-batch algorithm for different datasets. In the experiments, the set-batch algorithm creates 1,349, 2,079, 12,652, and 9,266 batches on JingDong, DC, TaFeng, and TaoBao, respectively. This can significantly improve the model efficiency since the model has to chronologically deal with 15,195, 42,905, 73,302, and 225,989 user-set interactions (see \tabref{tab:dataset_description}) without the set-batch algorithm. We report the running time of each epoch in the training and evaluation process in \tabref{tab:runtime_comparison}.

\begin{table}[!htbp]
\centering
\caption{Running time of CTTSP and CTTSP w/o set-batch on all the datasets.}
\label{tab:runtime_comparison}
\resizebox{1.01\columnwidth}{!}
{
\setlength{\tabcolsep}{1.0mm}
{
\begin{tabular}{c|c|cccc}
\hline
Process                      & Methods             & JingDong & DC   & TaFeng & TaoBao \\ \hline
\multirow{4}{*}{Training}   & CTTSP               & 47s      & 109s & 400s   & 426s   \\
                            & CTTSP w/o set-batch & 143s     & 399s & 1007s  & 2050s  \\ \cline{2-6} 
                            & Speedups            & 3.04     & 3.66 & 2.52   & 4.81   \\ \cline{2-6} 
                            & Average             & \multicolumn{4}{c}{3.51}          \\ \hline
\multirow{4}{*}{Evaluation} & CTTSP               & 18s      & 45s  & 183s   & 235s   \\
                            & CTTSP w/o set-batch & 54s      & 152s & 331s   & 933s   \\ \cline{2-6} 
                            & Speedups            & 3.00     & 3.38 & 1.81   & 3.97   \\ \cline{2-6} 
                            & Average             & \multicolumn{4}{c}{3.04}          \\ \hline
\end{tabular}
}
}
\end{table}

From \tabref{tab:runtime_comparison}, we could observe that the proposed set-batch algorithm enables our CTTSP to be efficiently executed in a mini-batch manner and achieves 3.5$\times$ speedups in training and 3.0$\times$ speedups in evaluation on average. We also compare the convergence of CTTSP and CTTSP w/o set-batch. The results show that the set-batch algorithm leads to faster and more stable convergence since it enables CTTSP to be trained in a mini-batch manner rather than processing interactions one by one, which can reduce the variance of the parameter updates \cite{DBLP:journals/corr/Ruder16}.

\begin{figure}[!htbp]
    \centering
    \includegraphics[width=1.0\columnwidth]{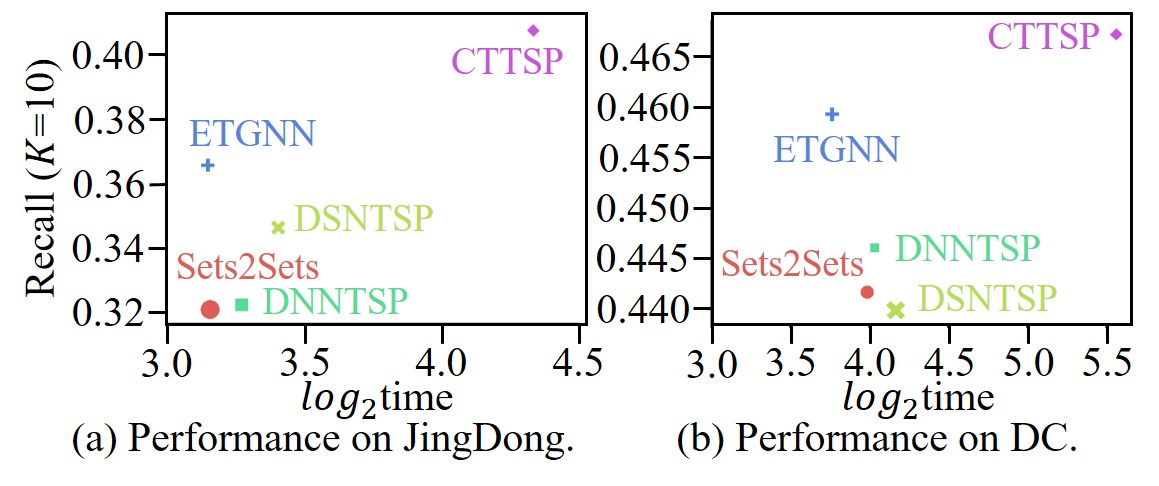}
    \caption{Log-scale evaluation time of different methods on JingDong and DC. The size of each point stands for the model parameter capacity.}
    \label{fig:runtime_comparison_JingDong_DC}
\end{figure}

We further report the evaluation time of different methods on JingDong and DC in \figref{fig:runtime_comparison_JingDong_DC}, where the x-axis is scaled by a logarithmic function with base 2. The results on all the datasets are shown in \appendixref{section-appendix-runtime-comparison}. The model complexity of CTTSP comes from the Continuous-Time User Preference Modelling and Personalized User Behavior Learning components, which is relatively higher than the baselines. However, we also observe that CTTSP obtains the best performance with acceptable increments in computational complexity. Therefore, we conclude that CTTSP can achieve a good trade-off between efficiency and effectiveness.

\subsection{Model Interpretability}
We show the interpretability of our approach by conducting experiments on the JingDong dataset.

\subsubsection{Element Representation Visualization}
\begin{figure}[!htbp]
    \centering
    \includegraphics[width=1.0\columnwidth]{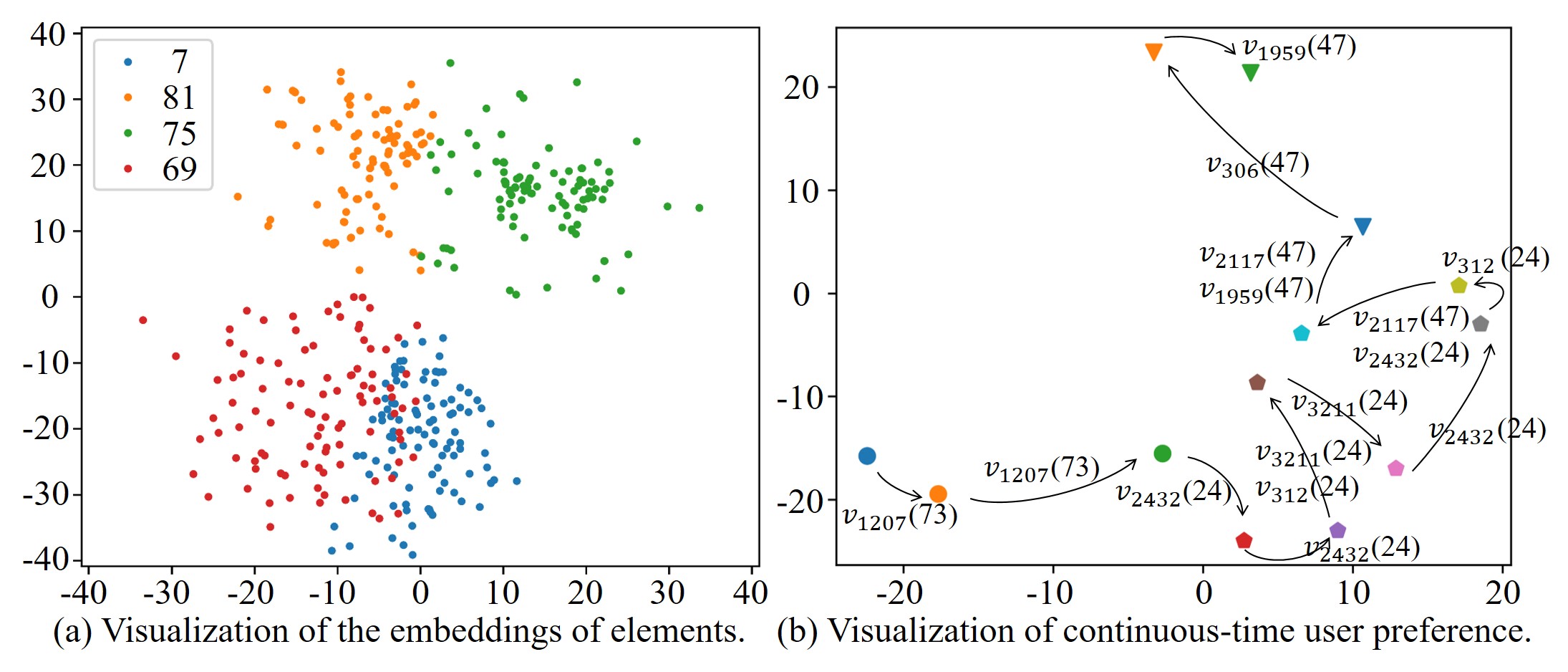}
    \caption{Analysis of the interpretability of our model.}
    \label{fig:model_interpretability}
\end{figure}

We first choose four categories that have the top four numbers of elements, and set the number of elements in each category to 100, leading to 400 elements in total. Then, we visualize the representations of the selected elements based on t-SNE \cite{van2008visualizing}. The results are shown in \figref{fig:model_interpretability}(a).

From \figref{fig:model_interpretability}(a), we could find that elements in the same category are closely gathered by our approach, and the boundaries between elements in different categories are relatively clear. This observation demonstrates the effectiveness of our method in capturing implicit correlations of elements. Note that our model does not explicitly access the category information of elements, but it still well learns such information. We also note that a few elements in \textit{category 7} and \textit{category 69} are mixed with each other, which may indicate that their underlying relationships can not be sufficiently represented by the category information only.

\subsubsection{Continuous-Time User Preference Visualization}
To confirm that our approach can track continuous-time user preferences by learning evolving memories of users, we first obtain all the memories of a sampled user, and then use t-SNE \cite{van2008visualizing} to project the memories into a two-dimensional space in \figref{fig:model_interpretability}(b). Each point represents the projected memory of the user after he/she interacts with a set at a specific time, and different points can reflect the time-evolving user preferences. The point shapes are determined by the types of elements in the interacted set. We chronologically connect the points with arrows based on the interaction time, aiming to show the changes in user preferences along the time dimension. The involved elements and their categories are also shown near the arrows.

From \figref{fig:model_interpretability}(b), we observe that the user's preference is evolving over time with different categories of the interacted elements. At first, the user displays a favor of element $v_{1207}$ in \textit{category 73}. Then, the user interacts with elements $v_{2432}$, $v_{312}$ and $v_{3211}$ in \textit{category 24}, and the user's preference gradually transits to the bottom-right corner of the latent space. Finally, the user's preference shifts to the top-right positions of the latent space because he/she shows interests in elements $v_{2117}$, $v_{1959}$ and $v_{306}$ in \textit{category 47}. These observations indicate that our approach can well capture the user's evolutionary preference by learning the continuously evolving memories.

\subsubsection{Collaborative Signal Exploration}
We present an intuitive case to show the ability of our CTTSP in exploring the collaborative signals latent in different users. In particular, we choose the sequences of three users including $u_{93}$, $u_{179}$ and $u_{950}$, and our task is to predict $u_{950}$'s next-period set, which contains a new element $v_{104}$ that $u_{950}$ has never interacted with.

\begin{figure}[!htbp]
    \centering
    \includegraphics[width=1.0\columnwidth]{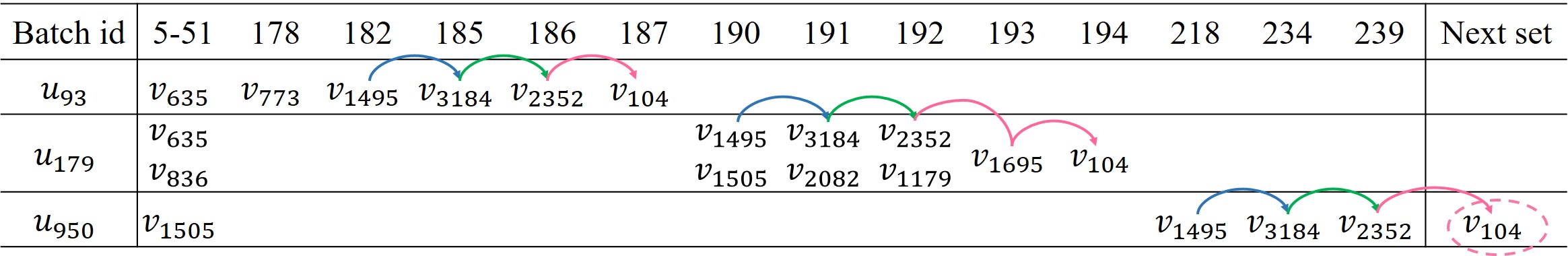}
    \caption{Analysis of how the learned collaborative signals benefit the prediction of user $u_{950}$'s next-period set.}
    \label{fig:collaborative_signals}
\end{figure}

The sequences of $u_{93}$, $u_{179}$ and $u_{950}$ are presented in \figref{fig:collaborative_signals}. We find that $u_{93}$ and $u_{179}$ show similar behaviors to user $u_{950}$, and we mark the element transaction patterns (i.e., underlying collaborative signals) with different colors. It would be helpful to leverage the collaborative signals when predicting the next-period set for user $u_{950}$. By chronologically learning from the constructed universal sequence that consists of all the users' historical behaviors, the proposed CTTSP first discovers the collaborative signals latent in user $u_{93}$ and user $u_{179}$, and then successfully predicts element $v_{104}$ for user $u_{950}$ within the top-5 results, while the baselines fail to do so. This observation demonstrates the superiority of our CTTSP in exploring collaborative signals for improving prediction performance.

\section{Conclusion}
\label{section-6}
In this paper, we studied the temporal sets prediction problem and pointed out the limitations of the existing methods in the implicit learning paradigm of user preferences. 
We proposed a continuous-time learning framework to explicitly capture the evolving user preferences by maintaining a memory bank, which could store the states of all the users and elements. We first constructed a non-descending universal sequence to contain all the user-set interactions, and then chronologically learned from each interaction. Our approach could update the memories of the user and elements for each interaction by message encoders and memory updaters.
A personalized user behavior learning module was also devised to capture each user's individual properties by adaptively aggregating the historical sequence.
Moreover, a set-batch algorithm was developed to improve the model efficiency.
Extensive experiments on four benchmarks showed that our approach significantly outperformed the existing baselines and revealed good interpretability.




%

%

\ifCLASSOPTIONcompsoc
  \section*{Acknowledgments}
\else
  \section*{Acknowledgment}
\fi
This work was supported by the National Natural Science Foundation of China (No. 62272023 and 51991395) and the Fundamental Research Funds for the Central Universities (No. YWF-23-L-1203).

\ifCLASSOPTIONcaptionsoff
  \newpage
\fi



\bibliographystyle{IEEEtran}
\bibliography{reference}

%

\begin{IEEEbiography}[{\includegraphics[width=1in,height=1.25in,clip,keepaspectratio]{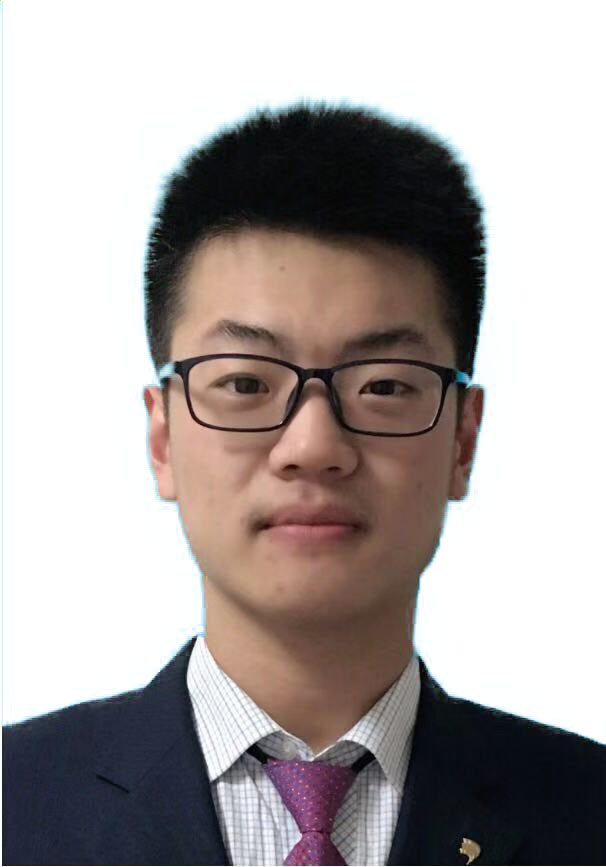}}]
{Le Yu} received the B.S. degree in the School of Computer Science and Engineering at Beihang University, Beijing, China, in 2019. He is currently a fifth-year computer science Ph.D. student in the School of Computer Science and Engineering at Beihang University. His research interests include temporal data mining, user behavior modeling, and graph neural networks.
\end{IEEEbiography}

\begin{IEEEbiography}[{\includegraphics[width=1in,height=1.25in,clip,keepaspectratio]{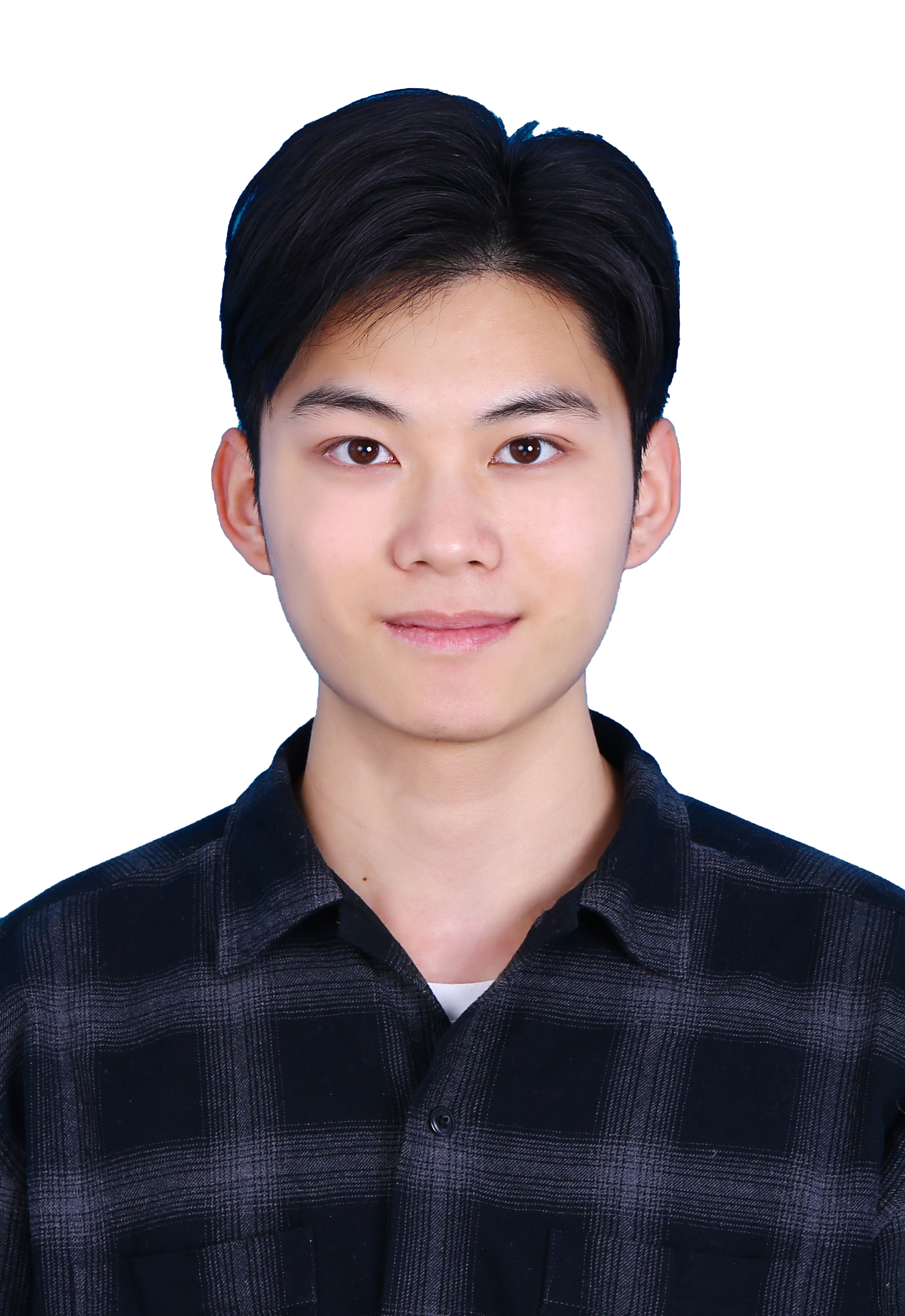}}]
{Zihang Liu} is currently a second-year M.S. student in Computer Science and Engineering from Beihang University. He received the B.S. degree in 2021. His research interests include time-series data mining, recommendation system, and graph neural networks.
\end{IEEEbiography}

\begin{IEEEbiography}[{\includegraphics[width=1in,height=1.25in,clip,keepaspectratio]{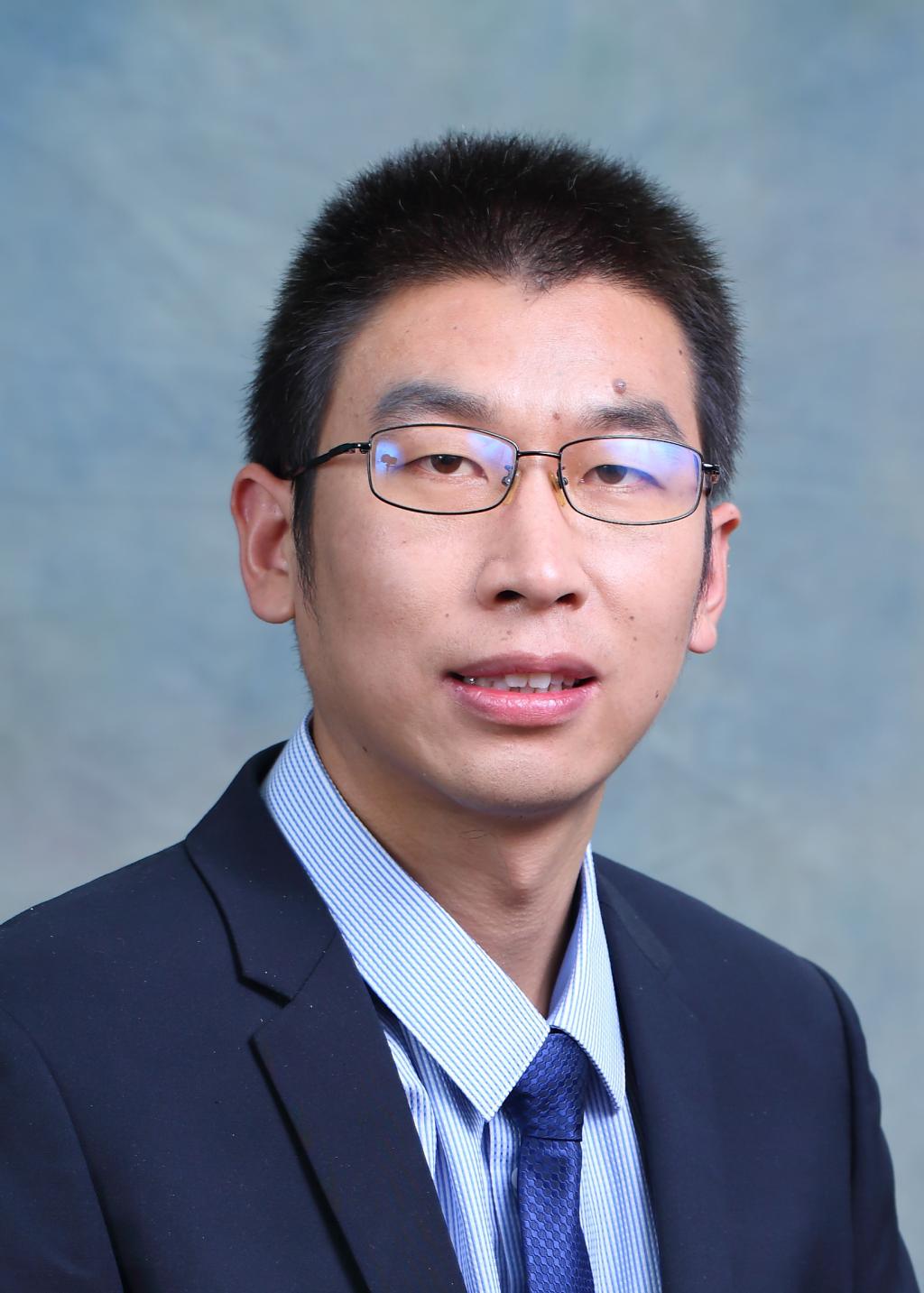}}]
{Leilei Sun} received the B.S., M.S. and Ph.D. degree from Dalian University of Technology in 2009, 2012, and 2017 respectively. He is currently an associate professor with the State Key Laboratory of Software Development Environment (SKLSDE) at School of Computer Science, Beihang University. He was a postdoctoral research fellow from 2017 to 2019 at Tsinghua University. His research interests include machine learning and data mining. 
\end{IEEEbiography}

\begin{IEEEbiography}[{\includegraphics[width=1in,height=1.25in,clip,keepaspectratio]{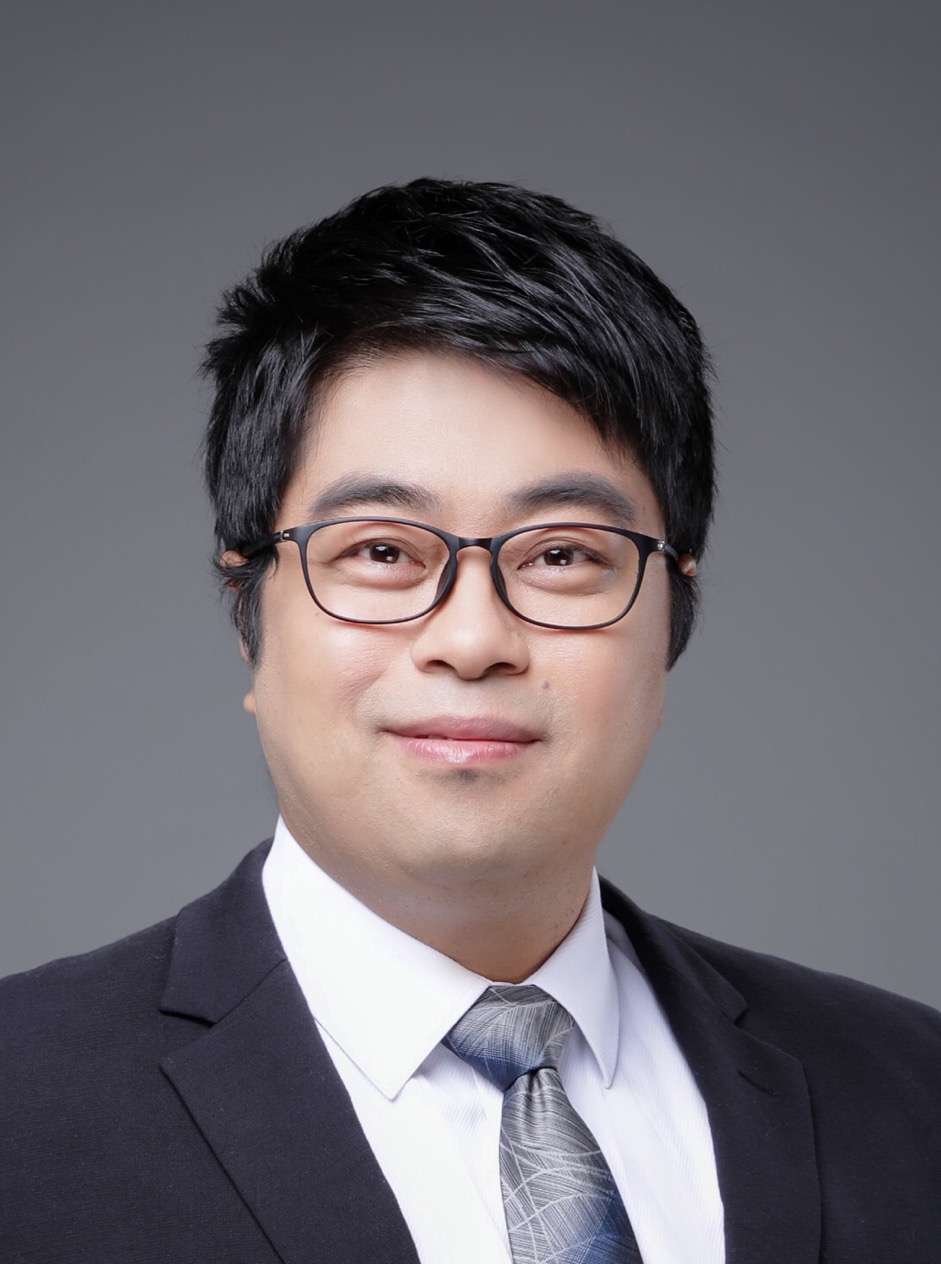}}]
{Bowen Du} received the Ph.D. degree in Computer Science and Engineering from Beihang University, Beijing, China, in 2013. He is currently a Professor with the State Key Laboratory of Software Development Environment, Beihang University. His research interests include smart city technology, multi-source data fusion, and traffic data mining.
\end{IEEEbiography}

\begin{IEEEbiography}[{\includegraphics[width=1in,height=1.25in,clip,keepaspectratio]{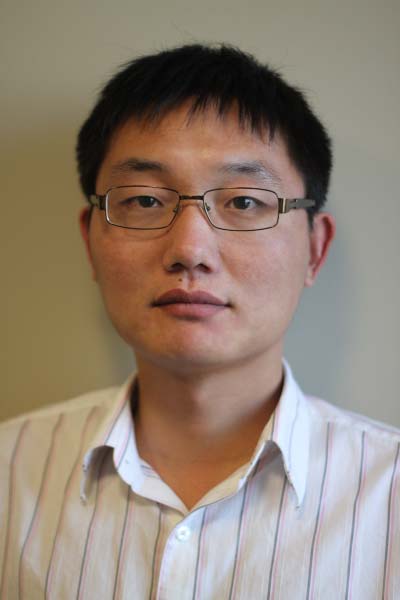}}]
{Chuanren Liu} received the B.S. degree from the University of Science and Technology of China (USTC), the M.S. degree from the Beijing University of Aeronautics and Astronautics (BUAA), and the Ph.D. degree from Rutgers, the State University of New Jersey.
He is currently an associate professor with the Business Analytics and Statistics Department at the University of Tennessee, Knoxville, USA.
His research interests include data mining and machine learning, and their applications in business analytics.
\end{IEEEbiography}


\begin{IEEEbiography}[{\includegraphics[width=1in,height=1.25in,clip,keepaspectratio]{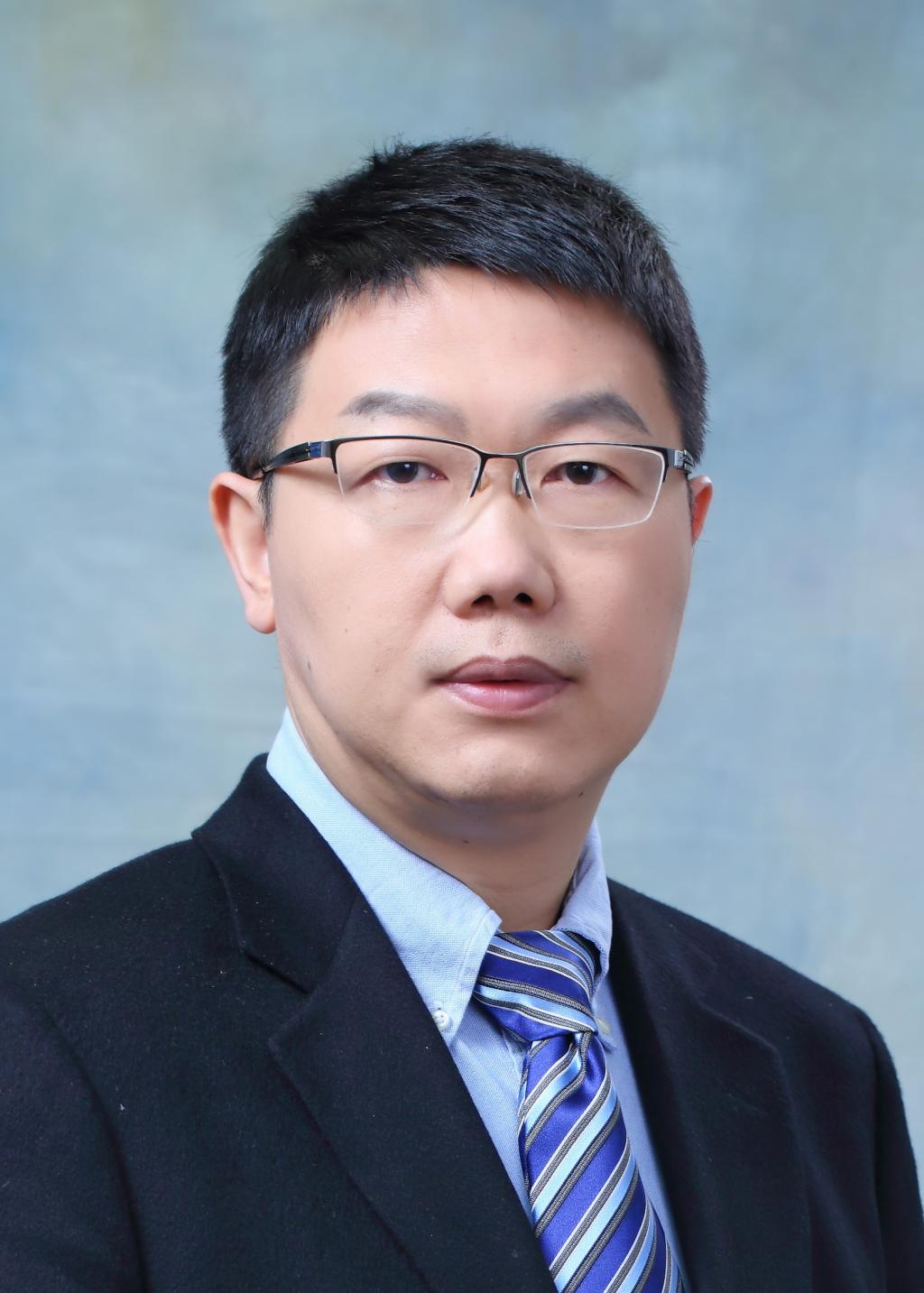}}]
{Weifeng Lv} received the B.S. degree in Computer Science and Engineering from Shandong University, Jinan, China, and the Ph.D. degree in Computer Science and Engineering from Beihang University, Beijing, China, in 1992 and 1998 respectively. Currently, he is a Professor with the State Key Laboratory of Software Development Environment, Beihang University, Beijing, China. His research interests include smart city technology and mass data processing.
\end{IEEEbiography}






\clearpage
\appendix
\label{section-appendix}
In the appendix, details of the experiments are introduced.

\subsection*{Statistics of Datasets} \label{section-appendix-data-statistics}
Statistics of the datasets are shown in \tabref{tab:dataset_description}, where \#E/S denotes the average number of elements in each set, and \#S/U stands for the average number of sets of each user.

\begin{table}[!htbp]
\centering
\caption{Statistics of the datasets. S, U, and E denote sets, users, and elements.}
\label{tab:dataset_description}
\resizebox{1.02\columnwidth}{!}
{
\setlength{\tabcolsep}{0.6mm}
{
\begin{tabular}{ccccccc}
\hline
Datasets & \#S  & \#U & \#E & \#E/S & \#S/U & Time span \\ \hline
JingDong & 15,195  & 3,063   & 3,551      & 1.26  & 4.96  & Mar. 1, 2018 -- Mar. 31, 2018 \\
DC       & 42,905  & 9,010   & 217        & 1.52  & 4.76  & the first 60 days \\
TaFeng   & 73,302  & 9,841   & 4,935      & 5.41  & 7.45  & Nov. 1, 2000 -- Feb. 28, 2001 \\
TaoBao   & 225,989 & 49,393  & 689        & 1.45  & 4.58  & Nov. 24, 2017 -- Dec. 3, 2017 \\ \hline
\end{tabular}
}
}
\end{table}

\subsection*{Settings of Hyperparameters} \label{section-appendix-hyper-parameters}
\tabref{tab:model_settings} shows the hyperparameter settings of our approach.

\begin{table}[!htb]
\centering
\caption{Hyperparameter settings of our approach on all the datasets.}
\label{tab:model_settings}
\begin{tabular}{c|cccc}
\hline
Settings  & JingDong   & DC & TaFeng & TaoBao   \\ \hline
Learning rate    & 0.001 & 0.001  & 0.001    & 0.001 \\
Dropout rate          & 0.2   &  0.2  & 0.15      & 0.05    \\
Hidden dimension $d$ & 64    & 64     & 64       & 32    \\
Hyperparameter $\lambda_{up}$ & 0.9   & 0.5    & 0.05     & 0.9    \\
Hyperparameter $\lambda_{cp}$ & 0.9   & 0.0    & 0.7     & 0.7     \\ \hline
\end{tabular}
\end{table}

\subsection*{Model Efficiency Comparison} \label{section-appendix-runtime-comparison}
\begin{figure}[!htbp]
    \centering
    \includegraphics[width=1.0\columnwidth]{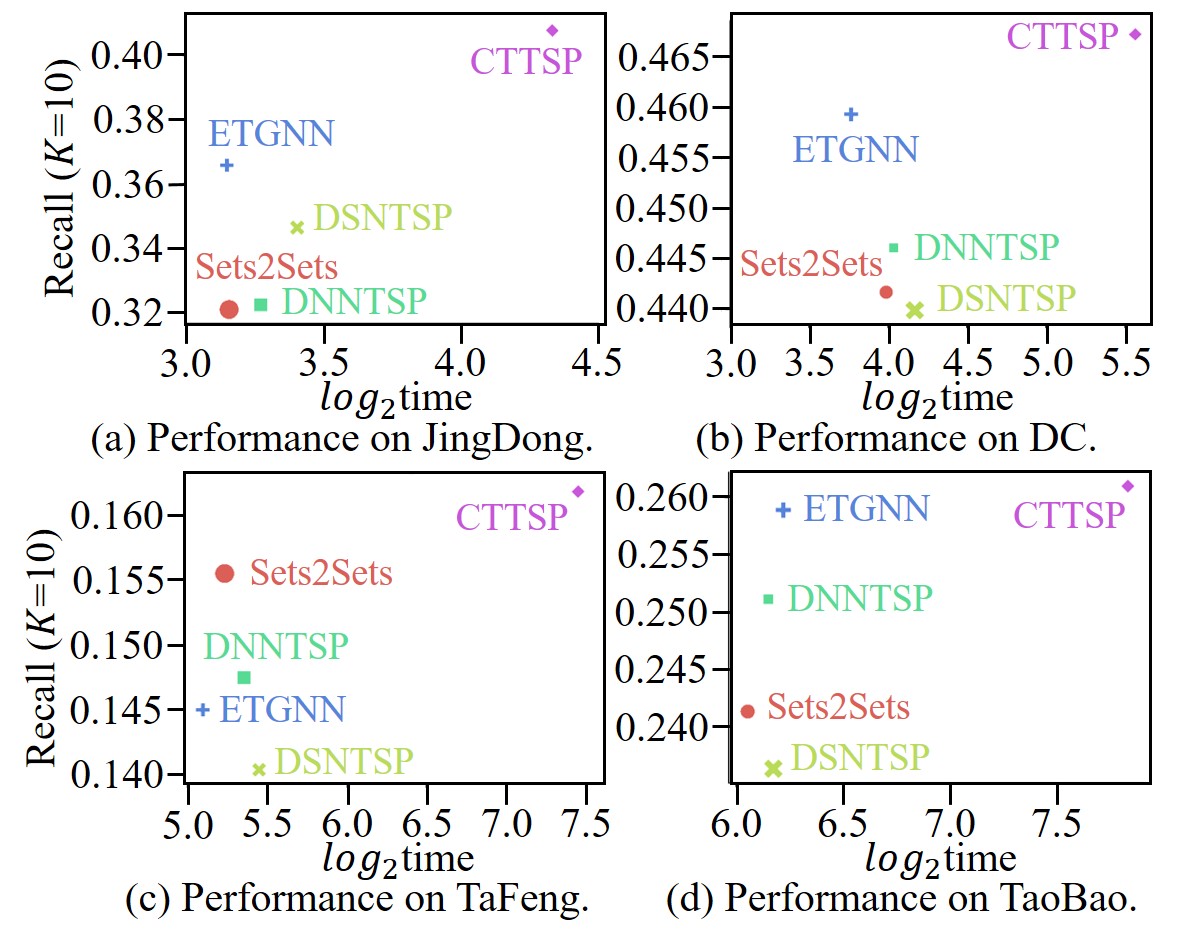}
    \caption{Log-scale evaluation time of different methods on all the datasets. The size of each point denotes the model parameter capacity.}
    \label{fig:runtime_comparison}
\end{figure}

\figref{fig:runtime_comparison} shows the evaluation time of different methods on all the datasets, where the x-axis is scaled by a logarithmic function with base 2.

\subsection*{Model Training Algorithm} \label{section-appendix-training-process}
The training process of CTTSP is shown in \algoref{alg:training_process}.

\begin{algorithm}[!htbp]
\SetKwComment{Comment}{/* }{ */}
\SetKwInOut{Input}{Input}
\SetKwInOut{Output}{Output}
\caption{Training process of CTTSP}
\label{alg:training_process}
\Input{Collection of users $\mathbb{U}$ and elements $\mathbb{V}$, a sequence of user-set interactions $\mathcal{S}^t=\{e_1,\cdots,e_K\}$ with $0 < t_1 \leq \cdots \leq t_K \leq t$, hyperparameters $\lambda_{up}$ and $\lambda_{cp}$, maximum number of training epochs $MaxEpoch$\;}
\Output{The model parameters $\Theta$ after training\;}
Initialize the parameters in CTTSP with random weights $\Theta$\;
Initialize the memories of all the users and elements in the memory bank with zero vectors\;
Create the batched data $\mathcal{B}=\left\{B_1,\cdots,B_{BatchNum}\right\}$ using the set-batch in \algoref{alg:set_batch}\;
$Epochs \gets 1$\;
\While{model is not converged and $Epochs \leq MaxEpoch$}{
    \For{batch $B_b \in \mathcal{B}$ }{
        $\{\bm{m}_i^{u,t_k}\}$ $\gets$ Encode messages of users in $B_b$ with $\{\bm{z}_i^{u,t_k^-(u_i)}\}$ and $\{\bm{z}_j^{v,t_k^-(v_j)}\}$ as inputs\;
        $\{\bm{m}_j^{v,t_k}\}$ $\gets$ Encode messages of elements in $B_b$ with $\{\bm{z}_i^{u,t_k^-(u_i)}\}$ and $\{\bm{z}_j^{v,t_k^-(v_j)}\}$ as inputs\;
        $\{\bm{z}_i^{u,t_k}\}$ $\gets$ Update memories of users in $B_b$ with $\{\bm{m}_i^{u,t_k}\}$ and $\{\bm{z}_i^{u,t_k^-(u_i)}\}$ as inputs\;
        $\{\bm{z}_j^{v,t_k}\}$ $\gets$ Update memories of elements in $B_b$ with $\{\bm{m}_j^{v,t_k}\}$ and $\{\bm{z}_j^{v,t_k^-(v_j)}\}$ as inputs\;
        $\{\bm{h}_{i,j}^{t_k}\}$ $\gets$ Aggregate the user and element perspectives with $\bm{e}^u$, $\{\bm{e}_j^v\}$ and $\lambda_{up}$ as inputs\;
        $\{\hat{y}_{i,j}^{t_k}\}$ $\gets$ Compute the probabilities by fusing continuous-time and personalized probabilities with $\{\bm{z}_i^{u,t_k}\}$, $\{\bm{z}_j^{v,t_k}\}$, $\{\bm{z}_j^{v,t_k^-(v_j)}\}$, $\{\bm{h}_{i,j}^{t_k}\}$, and $\lambda_{cp}$ as inputs\;
        Optimize the model parameters $\Theta$ by the back-propagation algorithm\;
    }
    $Epochs \gets Epochs + 1$\;
}
\end{algorithm}

\end{document}